\documentclass[letterpaper, 10 pt, conference, review]{ieeeconf}  

\IEEEoverridecommandlockouts                              

\usepackage{graphicx}
\usepackage{amsmath}
\usepackage{array}
\usepackage{url}
\title{\LARGE \bf
Towards Obstacle-Avoiding Control of Planar Snake Robots Exploring Neuro-Evolution of Augmenting Topologies
}

\author{Advik Sinha$^{1 \dagger}$, Akshay Arjun$^{2 \dagger}$, Abhijit Das$^{1}$ and Joyjit Mukherjee$^{2}$
\thanks{$^{1}$Authors are with the Department of Computer Science and Information Science, BITS Pilani Hyderabad Campus, Hyderabad, India.
        {\tt\small f20222004@hyderabad.bits-pilani.ac.in, abhijit.das@hyderabad.bits-pilani.ac.in}}%
\thanks{$^{2}$Authors are with the Department of Electrical and Electronics Engineering, BITS Pilani Hyderabad Campus, Hyderabad, India.
        {\tt\small f20220507@hyderabad.bits-pilani.ac.in, joyjit.mukherjee@hyderabad.bits-pilani.ac.in}}%
\thanks{$^{\dagger}$ Both authors have contributed equally to the work.}
}

\begin{document}

\maketitle
\thispagestyle{empty}
\pagestyle{empty}

\begin{abstract}

This work aims to develop a resource-efficient solution for obstacle-avoiding tracking control of a planar snake robot in a densely cluttered environment with obstacles. Particularly, Neuro-Evolution of Augmenting Topologies (NEAT) has been employed to generate dynamic gait parameters for the serpenoid gait function, which is implemented on the joint angles of the snake robot, thus controlling the robot on a desired dynamic path. NEAT is a single neural-network based evolutionary algorithm that is known to work extremely well when the input layer is of significantly higher dimension and the output layer is of a smaller size. For the planar snake robot, the input layer consists of the joint angles, link positions, head link position as well as obstacle positions in the vicinity. However, the output layer consists of only the frequency and offset angle of the serpenoid gait that control the speed and heading of the robot, respectively. Obstacle data from a LiDAR and the robot data from various sensors, along with the location of the end goal and time, are employed to parametrize a reward function that is maximized over iterations by selective propagation of superior neural networks. The implementation and experimental results showcase that the proposed approach is computationally efficient, especially for large environments with many obstacles. The proposed framework has been verified through a physics engine simulation study on PyBullet. The approach shows superior results to existing state-of-the-art methodologies and comparable results to the very recent CBRL\cite{Jia2021} approach with significantly lower computational overhead. The video of the simulation can be found here: https://sites.google.com/view/neatsnakerobot
\end{abstract}

\section{INTRODUCTION}

Snake robots are a class of highly articulated robotic systems inspired by the morphology and locomotion mechanisms of biological snakes. These robots consist of serially connected segments, each equipped with one or more actual degrees of freedom, enabling the system to achieve high levels of flexibility and redundancy \cite{liljeback2012,liljeback2013}. The control of planar snake robots involves managing a 2-dimensional output space through a high-dimensional input space, resulting in an over-actuated system often requiring a multi-layered control framework \cite{joyjit2021,Mukherjee2021book}. Traditionally, control of snake robots, such as obstacle avoidance has relied on empirically designed gaits, which are widely accepted \cite{hirose1972,hirose1990}. These gait functions although being quite effective in structured and non-undulated environments, often lack the flexibility and adaptability needed in unstructured or changing environments. Central pattern generator (CPG)-based methods have been explored to develop other kind of gait functions that may be suitable for different scenarios \cite{manzoor2019neural,wu2010}. However, these approaches fall short of providing a control or path planning framework for obstacle-rich environments.  

To achieve efficient motion in an obstacle-rich environment, various obstacle-aided control approaches have been explored \cite{transeth2008,yoshida2024}. Obstacle avoiding control schemes based on precise and accurate modeling \cite{tanaka2014,tanaka2017} or machine learning techniques \cite{ouyang2018} were also proposed for such environments. Reinforcement learning (RL) have been wide discussed and have shown promise in recent works \cite{liu2024_1,liu2024_2}. Bayesian controller design using RL has been employed on snake robots which have demonstrated significantly superior performance for obstacle avoidance \cite{Jia2021,Qu2023}. However, they often require carefully tuned network architectures and large amounts of training data, and may struggle in sparse-reward or non differentiable environments which is common in robotics. The proposed approach, on the other hand, provides a resource-efficient solution for the obstacle-avoiding control of a planar snake robot in an obstacle-dense environment.

NEAT (Neuro-Evolution of Augmenting Topologies) is an evolutionary algorithm that simultaneously evolves both the structure and weights of neural networks. Unlike gradient-based methods, it doesn't require differentiable models, making it effective in environments with sparse rewards, non-linear dynamics, or unknown optimal architectures. NEAT's ability to automatically discover compact, modular, and scalable control policies has made it a strong candidate for robotics, game AI, and industrial optimization \cite{10.1162/106365602320169811, patrascu2023, kimura2016}. Building on this foundation, we apply NEAT to evolve control policies for a snake robot. Previous work has demonstrated the ability of NEAT to evolve locomotion and coordination strategies in unstructured environments \cite{Shrestha2025}. Our work extends this by integrating local sensory inputs and pose information, enabling emergent adaptive locomotion across complex terrains. Using NEAT's topology evolution, we avoid manual tuning and allow coordinated behaviors to arise organically through evolution.

Our approach leverages 2D LiDAR sensors along with the pose and orientation of the snake robot’s head link to enable adaptive and efficient obstacle avoidance. The local environment is interpreted through LiDAR scans, while the head link’s kinematic state provides a reference for direction and alignment. These inputs are fed into a neural network evolved using NEAT, which iteratively optimizes both the weights and the architecture of the network over generations. NEAT begins with simple networks and gradually complexifies their topology through evolutionary operations such as mutation and crossover, allowing the system to autonomously discover effective control strategies tailored to the robot's morphology and sensing capabilities. The evolved network outputs two key variables that modulate the frequency and offset of a gait function, enabling smooth and responsive locomotion. This approach, implemented in the PyBullet physics simulator, allows the snake robot to adaptively navigate cluttered environments without relying on manually engineered behaviors or fixed network structures. The contributions of this work are summarized as:
\begin{itemize}
    \item NEAT-based computationally efficient approach has been proposed for solving the obstacle avoidance control for the planar snake robot;
    \item A custom reward function is proposed to ensure efficient navigation of the robot towards the goal while avoiding collisions with obstacles;
    \item Training and simulation of the proposed approach have been carried out in a PyBullet environment to demonstrate the efficacy of the proposed framework.
\end{itemize}


\section{Problem Statement} \label{sec:prob_state}
In this work, the control and navigation of a snake robot in an obstacle dense environment has been studied.
\subsection{System Description}
\begin{figure}[t]
\centering
\includegraphics[width=3.2in]{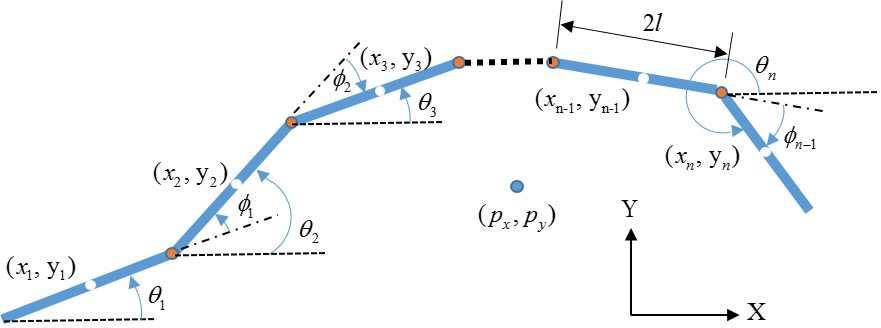}
\caption{Snake Robot Schematic}
\label{fig_schm}
\end{figure}

A snake robot is characterized by its modular, articulated overactuated structure, enabling high flexibility and redundant motion characteristics. Propagation is achieved through the coordinated body shape of robot links to generate traveling wave-like motions along the body, mimicking biological snakes. A schematic of a planar snake robot given in Fig. \ref{fig_schm} shows the articulated nature of the robot with $n$ number of links connected through $(n-1)$ actuated joints. The planar motion employs lateral undulation mode as a propagation, i.e., employing the serpenoid gait function to enforce time-varying undulations in the body to generate propulsive force through anisotropic friction. The gait function ensures a time-varying body shape of the snake robot which is instrumental in generating the necessary propulsive force. The serpenoid gait function \cite{hirose1972} for the \textit{i}th joint angle of the snake robot is defined as:

\begin{align}
\label{gait_eq}
\phi_{i} =
\begin{cases}
    \alpha \sin\left(\omega t + (i-1) \delta \right)&\text{for}~i=1,\ldots,(n-1),\\
    \alpha \sin\left(\omega t + (i-1) \delta \right) + \phi_0&\text{for}~i=n,
\end{cases}
\end{align}
where $\alpha$ is the amplitude, $\omega$ is the frequency, ${\delta}$ is the delay between each joint and $\phi_0$ is the offset. The value of $\alpha$ and ${\delta}$ are chosen as $0.6~rad$ and $\frac{\pi}{6}~rad$ respectively for our work. The frequency of the gait function $\omega$ determines the speed and the offset $\phi_0$ controls the direction of motion of the robot. Hence, by manipulating these two parameters, the overall trajectory of the snake robot can be controlled. 

\subsection{Objective}
An articulated system like snake robot is an overactuated system which provides flexible motion characteristics. However, the length of the robot also poses a challenge in cluttered environments where the density of obstacles are significantly high. As the length of the robot increases, the probability of collision also increases, and in some cases collisions are unavoidable \cite{Jia2021}. Although, the dynamics of a snake robot always allows a possible path, but seldom would have a collision free solution. Hence, right choice of objective function which minimizes the time to reach the goal as well as the number of collisions is of prime importance for the identification of a suitable path.

The serpenoid gait function provides a structured approach to control the joint angles by varying only two parameters $\omega$ and $\phi_0$ to obtain a desired body shape irrespective of the number of joints. This allows one to only find $2$ control parameters rather than $n-1$ control inputs making it computationally effective. However, it becomes difficult to assign or choose these gait parameters using classic means that would track a particular desired path, especially in an unstructured obstacle dense environment. Hence, NEAT based approach has been employed to obtain the desired gait parameters that could minimize the objective function while tracking a path towards the goal. The following sections will provide preliminary idea about NEAT as well as the application of the same for the obstacle avoiding control of a planar snake robot.

\section{Proposed methodology} \label{sec:prop_method}
This section provides a preliminary idea about NEAT, which has been employed in this work to solve the obstacle-avoiding control of a planar snake robot.
\subsection{Preliminaries}


NEAT \cite{10.1162/106365602320169811} is a genetic algorithm for the generation of evolving neural networks. Unlike traditional neuro-evolution approaches that evolve only the connection weights of a fixed neural architecture, NEAT evolves both the weights of the network, and the topology of the neural network. Each neural network, known as a genome, is represented as a set of genes that encodes the networks weights and structure. The NEAT algorithm primarily performs three steps: 
\begin{itemize}
    \item \textbf{Crossover between topologies} guided by \textit{innovation numbers}, which serve as historical markers assigned to genes at the time of their creation. These allow the algorithm to identify matching genes, which are genes in different genomes that originate from the same ancestral gene and therefore share the same innovation number. During crossover, matching genes are aligned and either averaged or randomly selected from one of the parents. Genes without a matching innovation number in the other parent are classified as either disjoint or excess. Disjoint genes occur within the range of innovation numbers common to both parents, whereas excess genes are outside that range. To preserve the useful mutations and promote innovation, NEAT inherits all disjoint and excess genes from the fitter parent, allowing structurally different but potentially advantageous traits to propagate.
    \item \textbf{Speciation}, which protects innovation and maintains diversity by grouping similar genomes into species using the compatibility distance function,
    \begin{equation}
    \delta = \frac{c_1 E}{N} + \frac{c_2 D}{N} + c_3 \cdot \bar{W},
    \end{equation} where E and D are the number of excess and disjoint genes, respectively. 
    ${\bar{W}}$ is the average weight difference of matching genes, N is the number of genes in the larger genome, and ${c_1}$, ${c_2}$, ${c_3}$ are coefficients. Two genomes are considered to be in the same species if their distance ${\delta}$ is below a predefined threshold ${\delta_t}$. By organizing genomes into species, NEAT ensures that individuals compete primarily within their own niche, thereby allowing new structural innovations to mature over generations without being prematurely eliminated by competition with more established architectures. This mechanism is especially important in NEAT, where topologies evolve dynamically alongside weights.
    \item \textbf{Incremental growth} is a core principle in NEAT, where neural networks begin with minimal topologies--typically just input and output nodes with no hidden layers and gradually grow in complexity over generations. This growth is driven by structural mutations that add new nodes by splitting existing connections, and new connections by linking previously unconnected nodes. The expansion allows NEAT to explore simple solutions first and only add complexity when beneficial. To preserve diversity and prevent premature convergence, NEAT employs fitness sharing within species. The raw fitness of each individual ${f_i}$ is first adjusted according to the density of its species:
    \begin{equation}
    f'_i = \frac{f_i}{\sum_{j=1}^{n} sh(\delta_d(i, j))},
    \end{equation} where the denominator sums up the sharing function ${sh(\delta_d)}$ over all ${n}$ individuals in the population. The sharing function is defined as:
    \begin{equation}
        sh(\delta_d) = 
    \begin{cases}
    1 & \text{if } \delta_d < \delta_t \\
    0 & \text{otherwise,}
    \end{cases}
    \end{equation}
    where $\delta_d$ is the compatibility distance between two genomes and ${\delta_t}$ is the compatibility threshold. Each species is then assigned a number of offspring in proportion to the sum of the adjusted fitness values of its members. Within a species, individuals compete to produce these offspring, ensuring that new innovations are protected while larger species cannot dominate solely due to size.
\end{itemize}

In the NEAT algorithm, each individual in the population represents a feedforward neural network (FNN) encoded as a genome. The genome defines the nodes (neurons) and connections (edges with weights), including information about which connections are enabled or disabled. Initially, networks have no hidden layers, but as evolution progresses, hidden nodes and new connections evolve via mutation.

The neural network computes outputs as a function of its inputs using the following operations: \begin{equation}
a_j = \sigma\left(\sum_{i \in \mathcal{I}(j)} w_{ij} \cdot a_i\right)
\end{equation}
where \(a_j\) is the activation of node \(j\), \(I(j)\) is the input to node \(j\), \(w_{ij}\) is the weight between node \(i\) and node \(j\), and \(\sigma(\cdot)\) is the activation function.

\subsection{Proposed NEAT-based approach for planar snake robot}
The proposed approach uses NEAT to evolve a neural network that serves as a controller for the snake robot. This neural network uses a set of simulated LiDAR values as the input providing information about the surrounding environment such as the obstacles and the target as well as internal robot parameters like joint angles. The network outputs two parameters, ${\omega}$ and ${\phi_0}$ of the gait function \eqref{gait_eq} which governs the movement of the snake robot. By adjusting the offset ${\phi_0}$, the neural network can control the curvature of the snake's body, allowing it to steer toward the target and avoid obstacles effectively. The speed of the robot also need to be manipulated to ensure the tracking of the desired obstacle avoiding path. The fitness function is designed to encourage both effective navigation toward the goal as well as collision avoidance. At each timestep, the reduction in distance along the goal direction (e.g. \textit{y}-axis towards the goal beacon) is rewarded as 
$\Delta d_y = d_y^{t-1} - d_y^{t}$ denotes the change in goal-axis distance between consecutive steps. A scaled reward is considered as
\begin{equation}
\label{f_prog_eq}
f_{\text{progress}} = \beta_p \max\!\left\{\Delta d_y, \,-\epsilon\right\},
\end{equation}
where $\beta_p$ is a gain factor and $\epsilon$ prevents excessively large negative penalties if the snake temporarily regresses. To discourage lateral drift away from the feasible path, a penalty is applied if the head position exceeds corridor bounds [$x_{min}$, $x_{max}$] as follows:

\begin{equation}
\label{f_corr_eq}
f_{\text{corridor}} = -\beta_c \max \left\{ 0, \, x_{\min} - x, \, x - x_{\max} \right\},
\end{equation}
where $\beta_c$ is the penalty for moving out of the defined corridor.
This corridor-based penalty arises naturally from the structure of the environment, where constraining the snake to a narrow lateral region encourages faster progress towards the goal. In more general environments, similar task-specific constraints can be defined to guide behavior. If a collision with an obstacle or the wall occurred during a step, a fixed collision penalty is applied,
\begin{equation}
\label{f_coll_eq}
f_{\text{collision}} = -\gamma \cdot \mathbf{1}_{\text{collision}} ,
\end{equation}

where ${\textbf{1}}_{\text{collision}}$ = 1 if a collision occurs in that step, and 0 otherwise, and $\gamma$ is the fixed collision penalty. Finally, to prevent premature stagnation and encourage exploration, a small step-wise bonus is added,
\begin{equation}
\label{f_live_eq}
f_{\text{living}} = \lambda
\end{equation}

The total fitness at each step is accumulated as:
\begin{equation}
\label{fitness_eq}
f = f_{\text{progress}} + f_{\text{corridor}} + f_{\text{collision}} + f_{\text{living}} 
\end{equation}

Additionally, a bonus is awarded for reaching the goal, and the simulation is terminated early upon multiple collisions. We train the NEAT algorithm in a complex, obstacle-dense environment to evolve a neural network controller for planar snake locomotion. The algorithm is trained over several generations to allow for effective learning and sufficient exploration of the solution space, with NEAT incrementally evolving both the weights and topologies of neural networks. Speciation preserves diversity by protecting novel topologies from premature extinction, while crossover and mutation operators introduce new structural and parametric variations that enable continuous adaptation to the task.

Unlike a simple open arena, the proposed environment is designed to be challenging: multiple layers of staggered cylindrical obstacles form a maze-like structure requiring coordinated turning manoeuvres. Solid walls further confine the arena, and a beacon placed at the far end serves as the goal. This setup requires the evolved controllers to demonstrate not only forward locomotion but also obstacle avoidance and path planning.

\begin{figure}[t]
\centering
\includegraphics[width=3in]{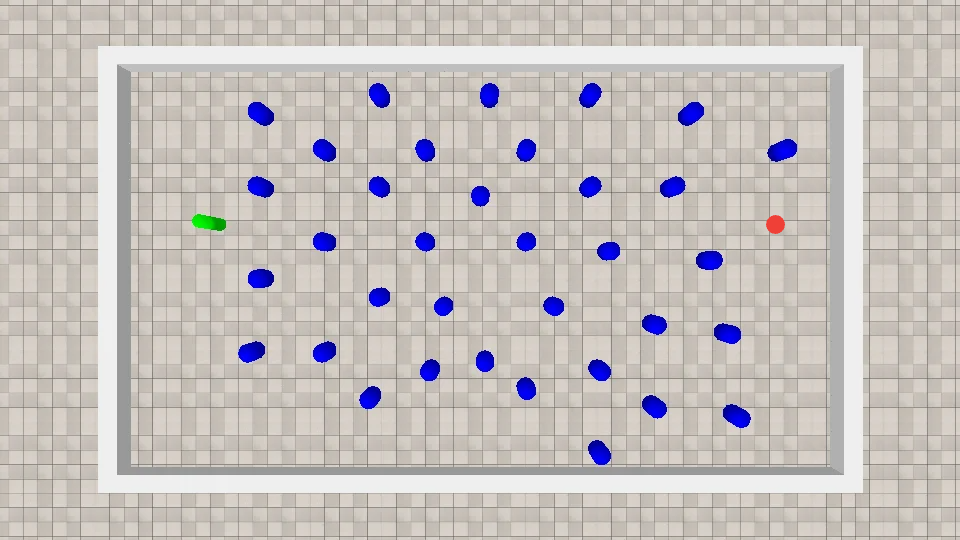}
\caption{Obstacle Dense Environment: The blue cylinders represent the obstacles, the white blocks are the arena walls, and the green cylinder represents the goal beacon. The red dot is the starting point of the snake}
\label{env_schm}
\end{figure}

\begin{table}[t]
\centering
\caption{SPECIFICATIONS OF THE SNAKE ROBOT USED}
\label{tab:specs}
\begin{tabular}{lcccc}
\hline
\textbf{Items} & \textbf{Values}\\
\hline
Joint Number & 8 \\
Link Size ($\mathrm{mm}^3$) & 100 x 50 x 50   \\
Link Weight (g) & 300    \\
Wheel Width (mm) &  10    \\
Wheel Radius (mm) & 30\\
Wheel Weight (g) & 10    \\
yaw angle range (deg) & [-45 , +45] \\
\hline
\end{tabular}
\end{table}

\section{Simulation and Experimental Validation} \label{sec:results}
This section details the training and simulation of the proposed methodology for the planar snake robot in a large obstacle dense environment.
\subsection{Simulation Scenario}
The simulation environment is built using PyBullet and confines the snake robot to a bounded 2-D rectangular arena, as shown in Fig. \ref{env_schm}. We replicate the environment used in CBRL \cite{Jia2021} as closely as possible. The arena is designed to impose both lateral constraints and obstacle avoidance objectives, compelling the robot to learn controlled and adaptive locomotion while progressing toward the goal beacon.

\begin{table}[!t]
\centering
\caption{NEAT Configuration Parameters}
\label{tab:neat_config}
\scriptsize
\begin{tabular}{|l|l|}
\hline
\textbf{Parameter} & \textbf{Value} \\
\hline
\texttt{fitness\_criterion} & max \\
\texttt{fitness\_threshold} & 1000 \\
\texttt{pop\_size} & 100 \\
\texttt{reset\_on\_extinction} & True \\
\texttt{activation\_default} & identity \\
\texttt{activation\_mutate\_rate} & 0.01 \\
\texttt{activation\_options} & identity \\
\texttt{aggregation\_default} & sum \\
\texttt{aggregation\_mutate\_rate} & 0.01 \\
\texttt{aggregation\_options} & sum \\
\texttt{bias\_init\_mean} & 0.0 \\
\texttt{bias\_init\_stdev} & 2.0 \\
\texttt{bias\_max\_value} & 30.0 \\
\texttt{bias\_min\_value} & -30.0 \\
\texttt{bias\_mutate\_power} & 1.5 \\
\texttt{bias\_mutate\_rate} & 0.7 \\
\texttt{bias\_replace\_rate} & 0.1 \\
\texttt{compatibility\_disjoint\_coefficient} & 1.0 \\
\texttt{compatibility\_weight\_coefficient} & 0.5 \\
\texttt{conn\_add\_prob} & 0.6 \\
\texttt{conn\_delete\_prob} & 0.3 \\
\texttt{enabled\_default} & True \\
\texttt{enabled\_mutate\_rate} & 0.01 \\
\texttt{feed\_forward} & True \\
\texttt{initial\_connection} & full\_direct \\
\texttt{node\_add\_prob} & 0.5 \\
\texttt{node\_delete\_prob} & 0.2 \\
\texttt{num\_hidden} & 40 \\
\texttt{num\_inputs} & 120 \\
\texttt{num\_outputs} & 2 \\
\texttt{response\_init\_mean} & 0.0 \\
\texttt{response\_init\_stdev} & 2.0 \\
\texttt{response\_max\_value} & 30.0 \\
\texttt{response\_min\_value} & -30.0 \\
\texttt{response\_mutate\_power} & 0.0 \\
\texttt{response\_mutate\_rate} & 0.0 \\
\texttt{response\_replace\_rate} & 0.0 \\
\texttt{weight\_init\_mean} & 0.0 \\
\texttt{weight\_init\_stdev} & 2.0 \\
\texttt{weight\_max\_value} & 30 \\
\texttt{weight\_min\_value} & -30 \\
\texttt{weight\_mutate\_power} & 3.0 \\
\texttt{weight\_mutate\_rate} & 0.8 \\
\texttt{weight\_replace\_rate} & 0.3 \\
\texttt{compatibility\_threshold} & 4.0 \\
\texttt{species\_fitness\_func} & max \\
\texttt{max\_stagnation} & 20 \\
\texttt{species\_elitism} & 2 \\
\texttt{elitism} & 2 \\
\texttt{survival\_threshold} & 0.3 \\
\hline
\end{tabular}
\end{table}

Obstacles are distributed in three categories:
\begin{itemize}
    \item \textbf{Perimeter walls} form the outer boundaries of the environment, enclosing the robot within a fixed exploration space. These are cubes with a side length of 0.5 unit.
    \item \textbf{Cylindrical obstacles} are placed throughout the arena in a maze-like fashion, forcing the snake robot to coordinate turns and navigate through narrow passages to reach the goal. These cylinders have a radius of 0.25 units and a length of 0.5 unit.
    \item A \textbf{goal beacon} is placed at the opposite end of the arena, providing a clear navigation objective for the snake. This is a cylinder with a radius of 0.15 units, and a length of 1 unit.
\end{itemize}

The physical parameters of the snake robot considered for the learning and simulation are given in Table \ref{tab:specs}. The snake spawns at $(1.5, 0.5)$, while the goal beacon is present at $(1.5, 16.5)$. It is to be noted, that the chances of finding a suitable path with minimum collisions for a given environment reduce drastically as the length of the link as well as the number of links increase.

\subsection{Implementation Details}
In our training framework, we use the PyBullet physics engine \cite{coumans2021}, which provides a simulation environment capable of handling complex dynamics, joint control, and accurate collision detection in real-time. The simulation uses a gravity-enabled, non-real-time physics model, with flat ground provided by a standard plane URDF. The step size is  For the learning component, we utilize the NEAT-Python library\cite{McIntyre_neat-python}, an implementation of the NEAT algorithm, which evolves both the weights and topologies of neural networks.

\begin{figure*}[t]
    \centering
    \begin{tabular}{cccc}
        \includegraphics[width=0.23\textwidth]{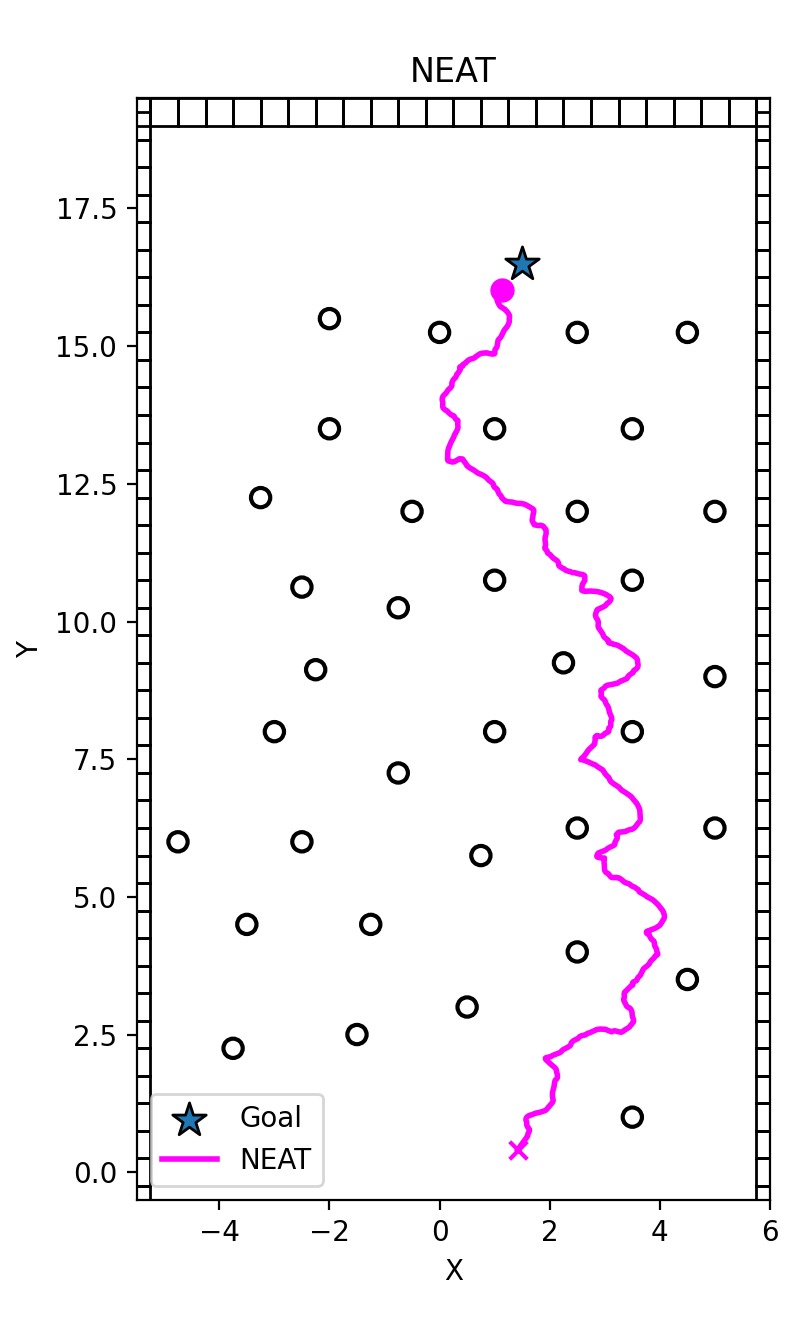} &
        \includegraphics[width=0.23\textwidth]{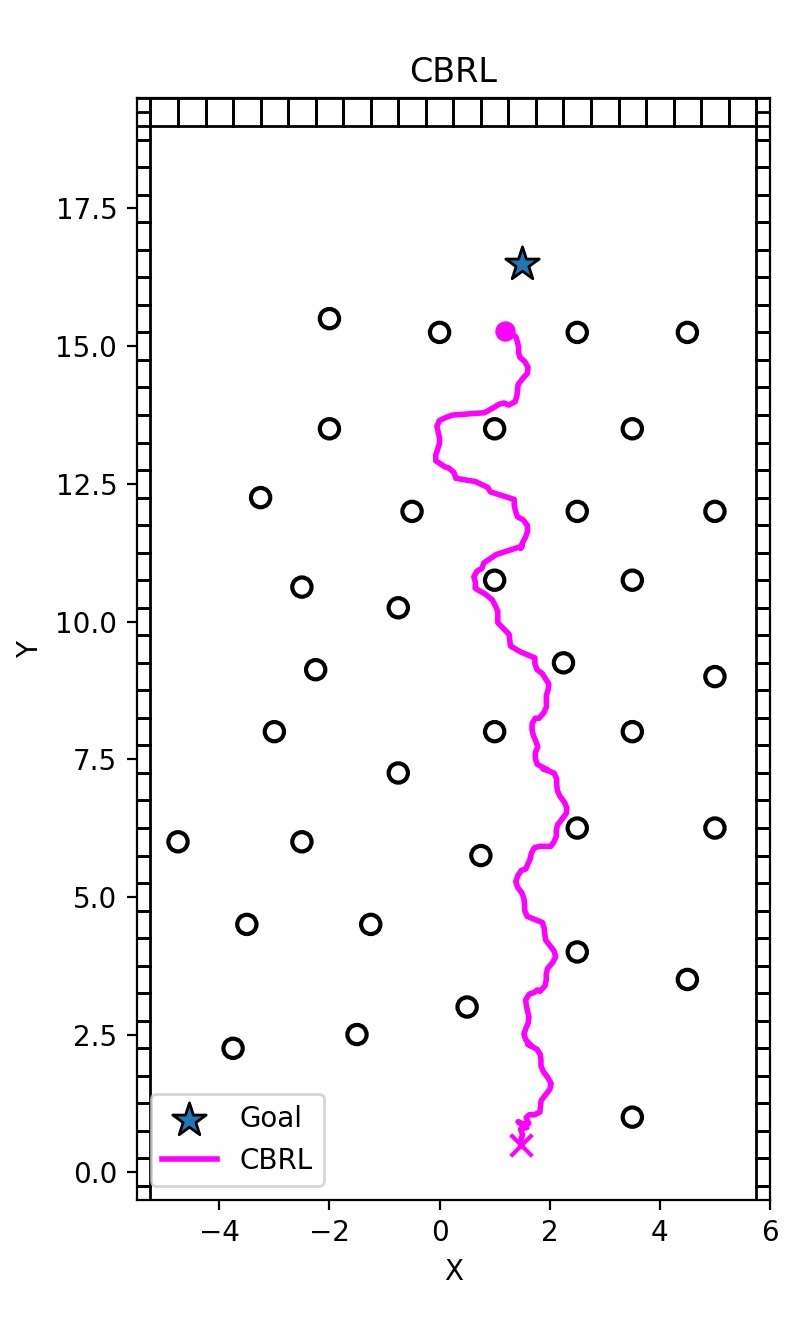} &
        \includegraphics[width=0.23\textwidth]{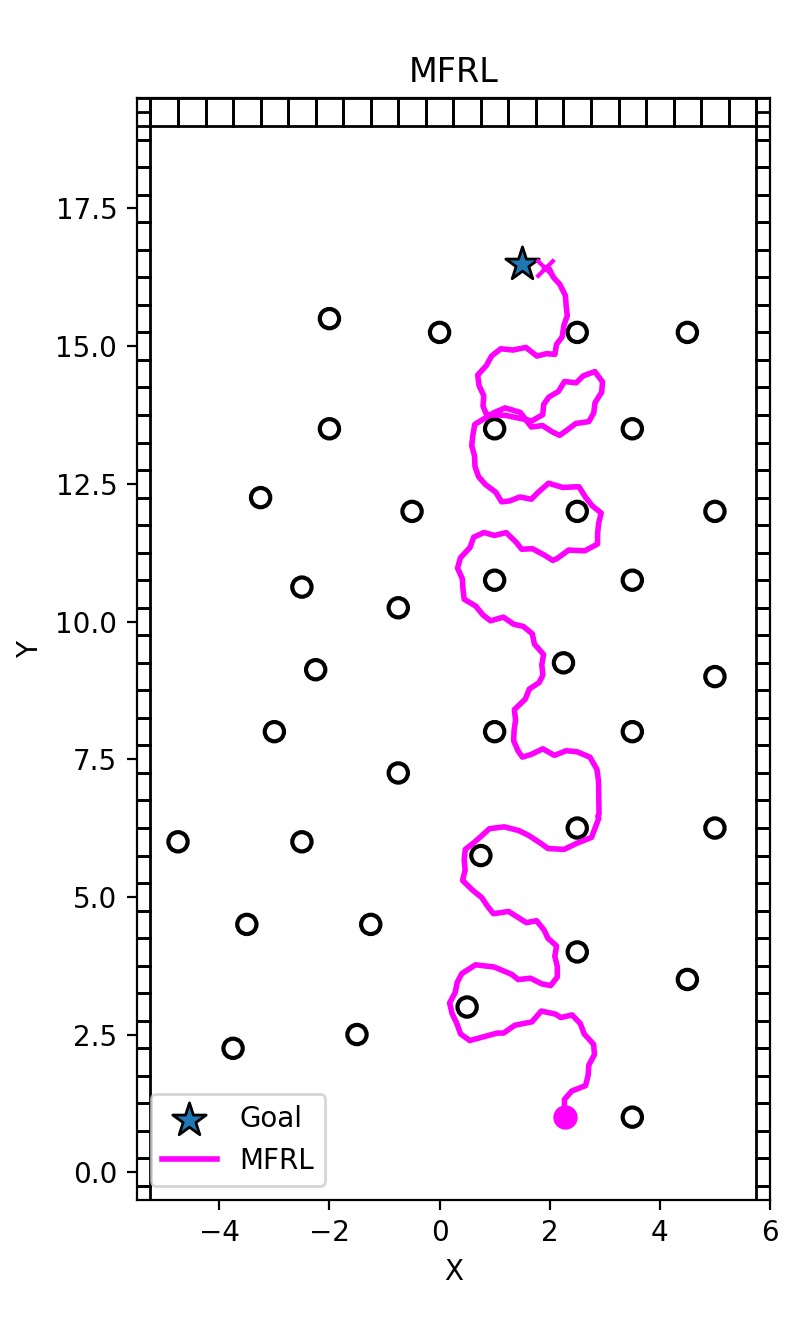} &
        \includegraphics[width=0.23\textwidth]{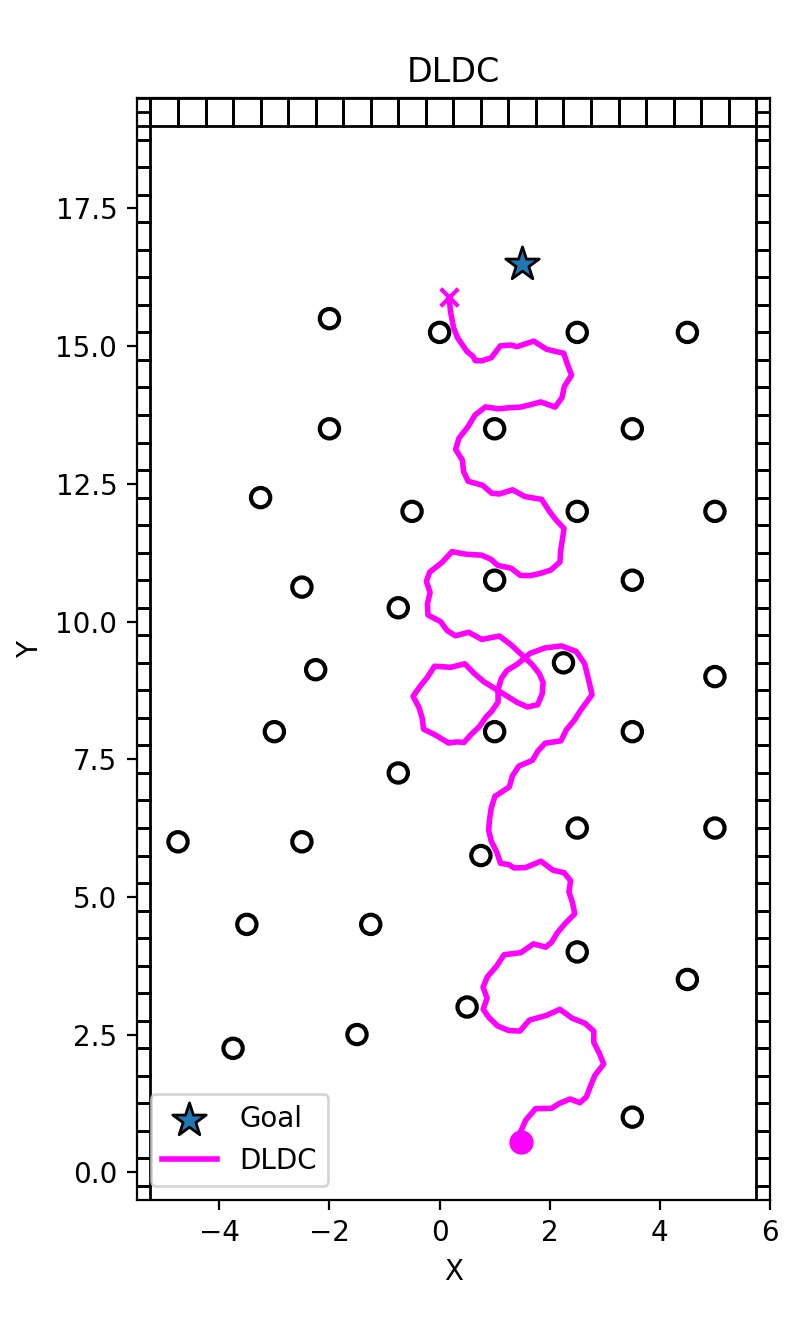} \\
    \end{tabular}
    \caption{Comparison of the path taken by the snake in NEAT, CBRL \cite{Jia2021}, MFRL \cite{8103164} and DLDC \cite{Sartoretti_2019}. The cross is the starting point, and the dot represents the end point of the snake.}
    \label{fig:prev results}
\end{figure*}

The configuration used for the NEAT algorithm is provided in Table \ref{tab:neat_config}. Every third LiDAR sensor value (from the 360 values) is sampled as input for the NEAT algorithm, thus giving us the 120 inputs required for the NEAT algorithm.
The first output of the neural network is used to compute the wave frequency parameter, ${\omega}$, 
for the gait equation (\ref{gait_eq}). This is done using a sigmoid activation as follows:
\begin{equation}
\label{omega}
{\omega} = a + b \cdot \frac{1}{1 + e^{-o_1 / \tau}},
\end{equation}
where ${o_1}$ is the first output of the neural network, ${a}$ and ${b}$ are hyperparameters, and $\tau$ is the temperature parameter controlling the sharpness of the sigmoid output. In our experiments, we set \( a = 0.5 \) and \( b = 2.0 \). The coefficient \( b \) serves as a \textit{scaling factor}, expanding the effective output range of the sigmoid to allow for more expressive frequency modulation. The offset \( a \) ensures that the base frequency starts from a non-zero minimum, preventing the robot from producing ineffective or static movement patterns at low activations. This formulation results in \( w \) ranging between 0.5 and 2.5.

The second output of the neural network, \( o_2 \), is passed through a hyperbolic tangent function to compute the parameter \( \phi_0 \) as follows:
\begin{equation}
\label{phi}
\phi_0 = \tanh(\frac{o_2}{\tau}),
\end{equation}
where $\tau$ is the temperature parameter controlling the sharpness of the \(\tanh\) activation.
This transformation ensures that \( \phi_0 \) remains bounded within the range \((-1, 1)\), which is important for controlling the robot's turning behaviour. Specifically, \( \phi_0 \) is added as an offset to the first servo joint in the snake, allowing the network to influence the direction of movement. The choice of the \(\tanh\) function over other activation functions provides smooth, continuous output and suppresses extreme values, which helps stabilize the motion and provides gradual changes in direction. In this work, the temperature, $\tau$ is set to $3$ for both (\ref{omega}) and (\ref{phi}).

\begin{figure*}[t]   
    \centering
    \begin{tabular}{ccc}
        \includegraphics[width=0.33\textwidth, height=4cm]{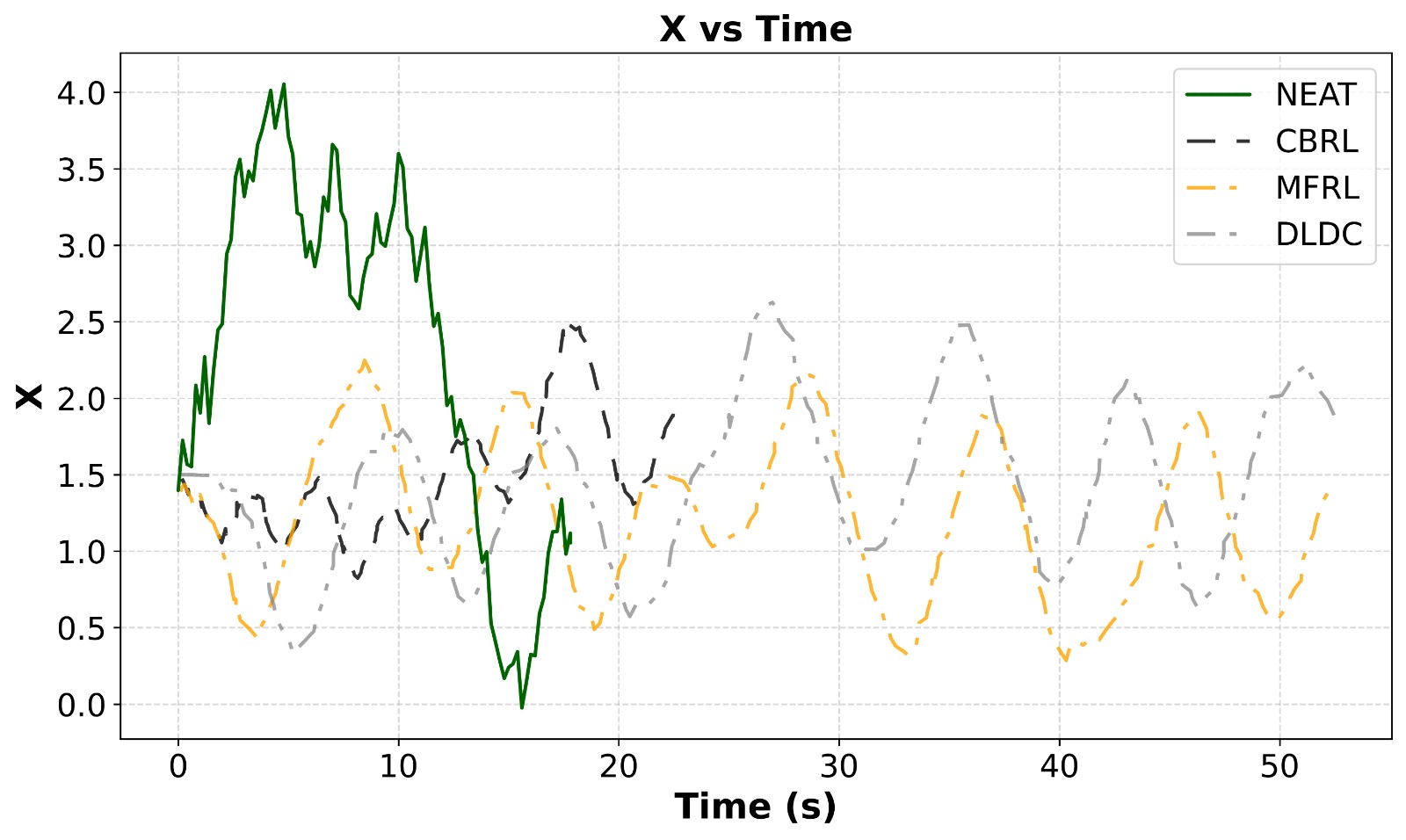} &
        \includegraphics[width=0.33\textwidth, height=4cm]{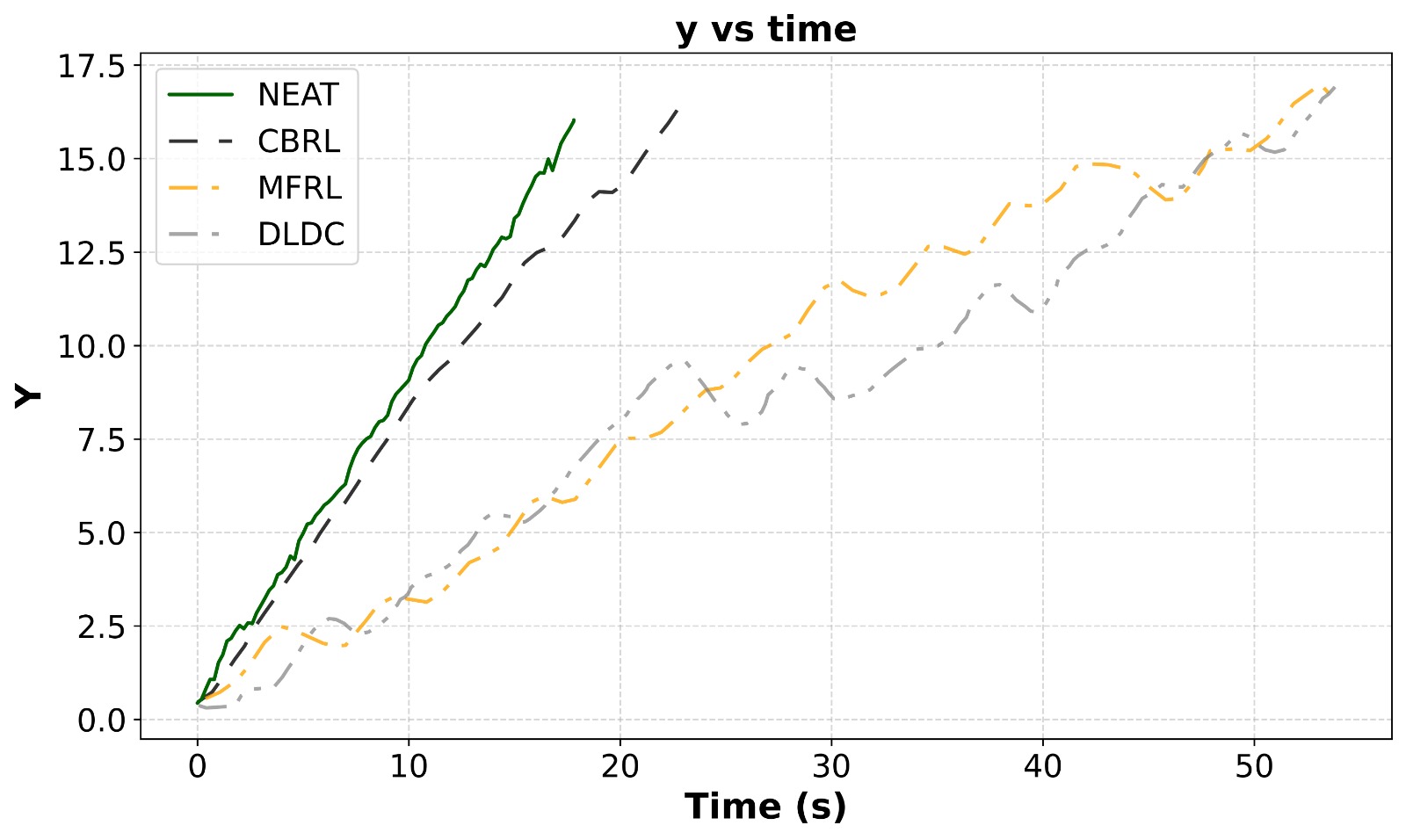} &
        \includegraphics[width=0.33\textwidth, height=4cm]{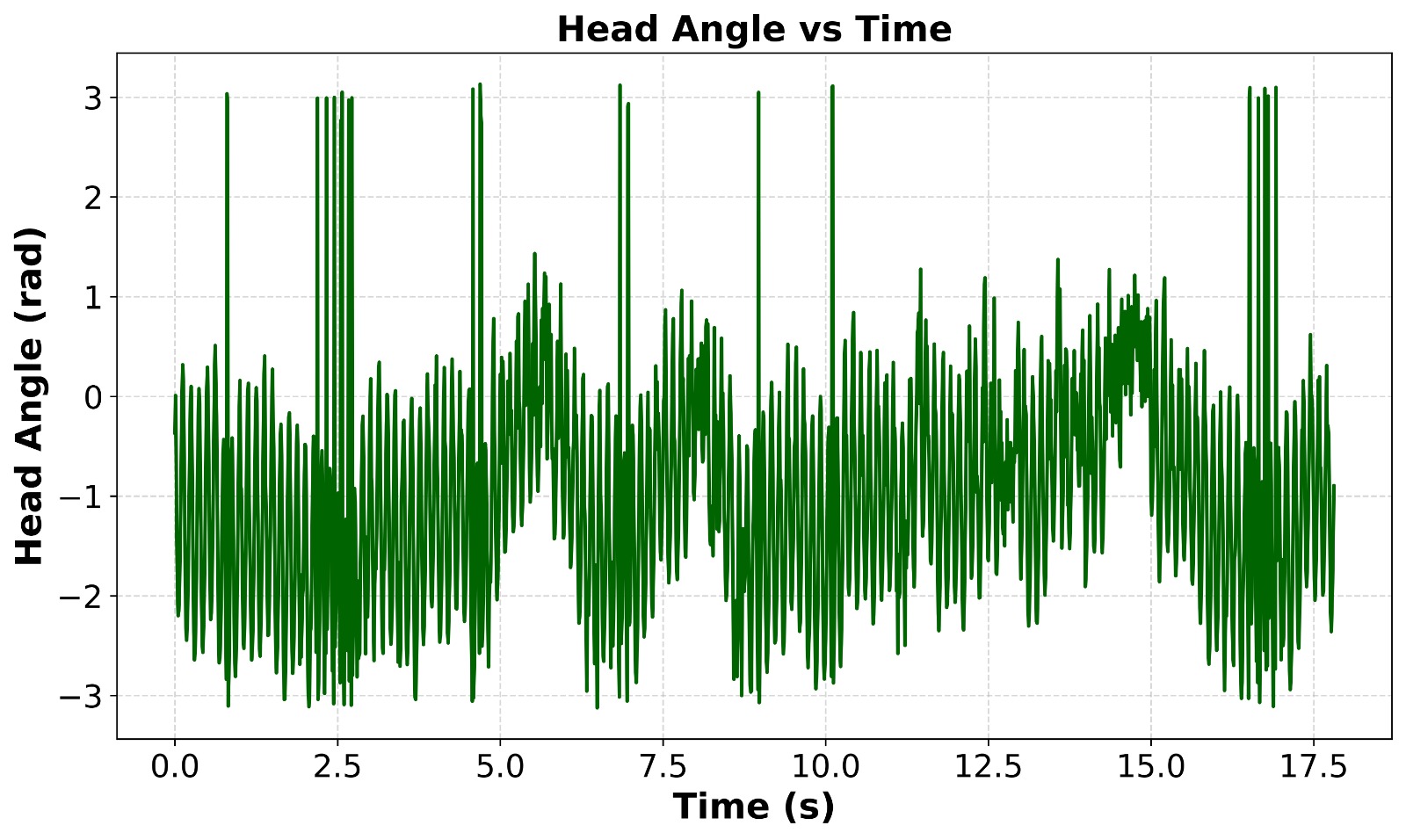} \\
        \includegraphics[width=0.33\textwidth, height=4cm]{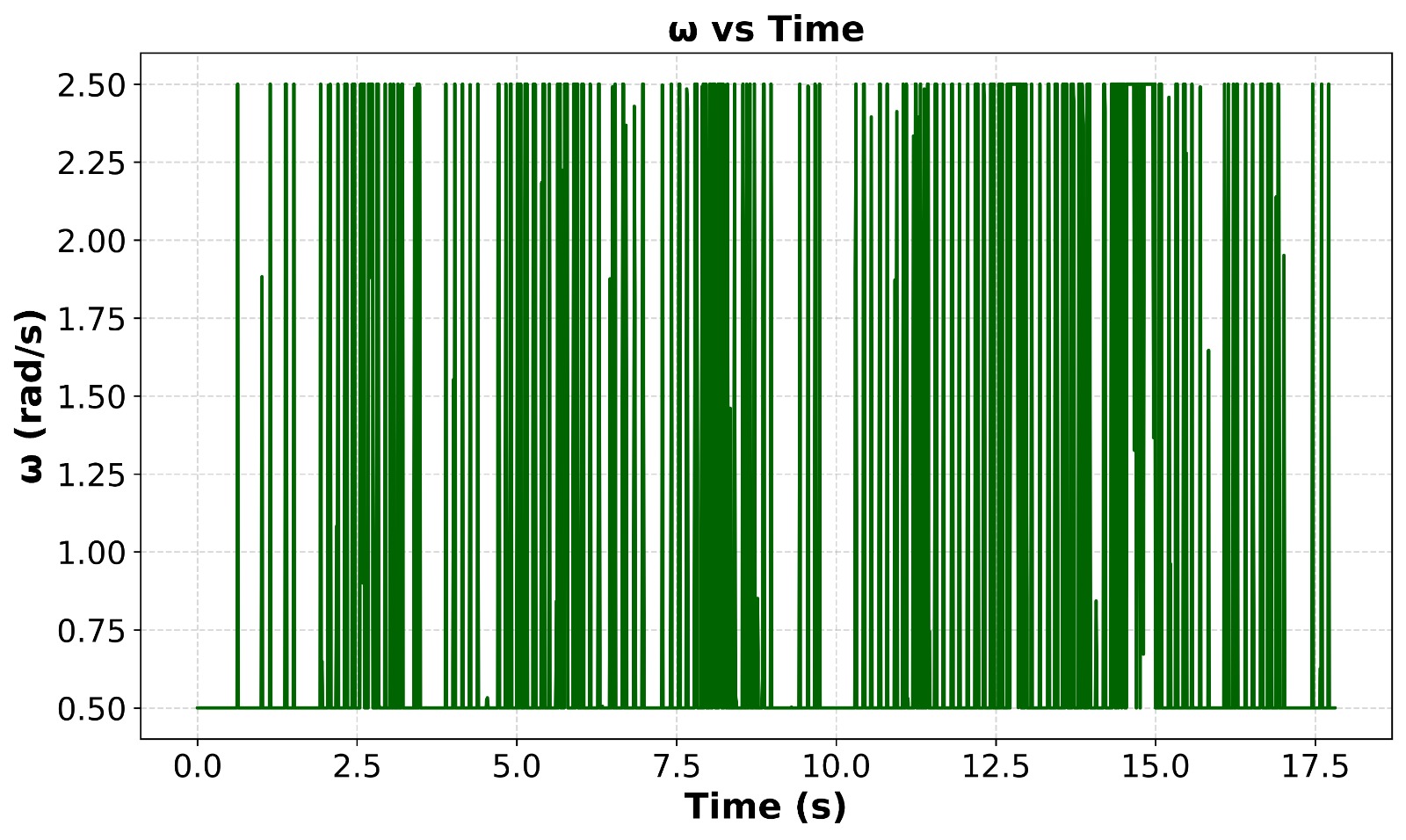} &
        \includegraphics[width=0.33\textwidth, height=4cm]{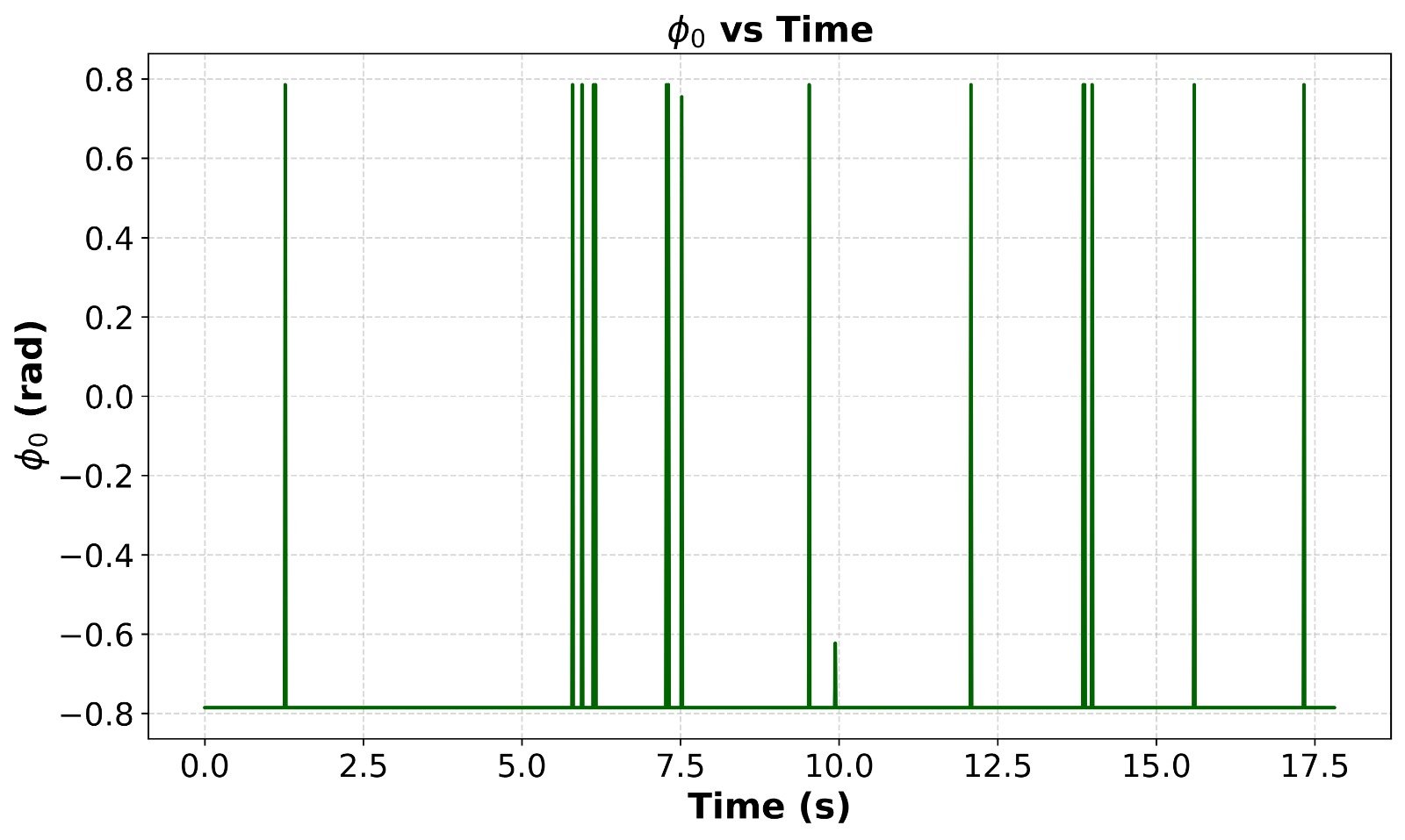} &
        \includegraphics[width=0.33\textwidth, height=4cm]{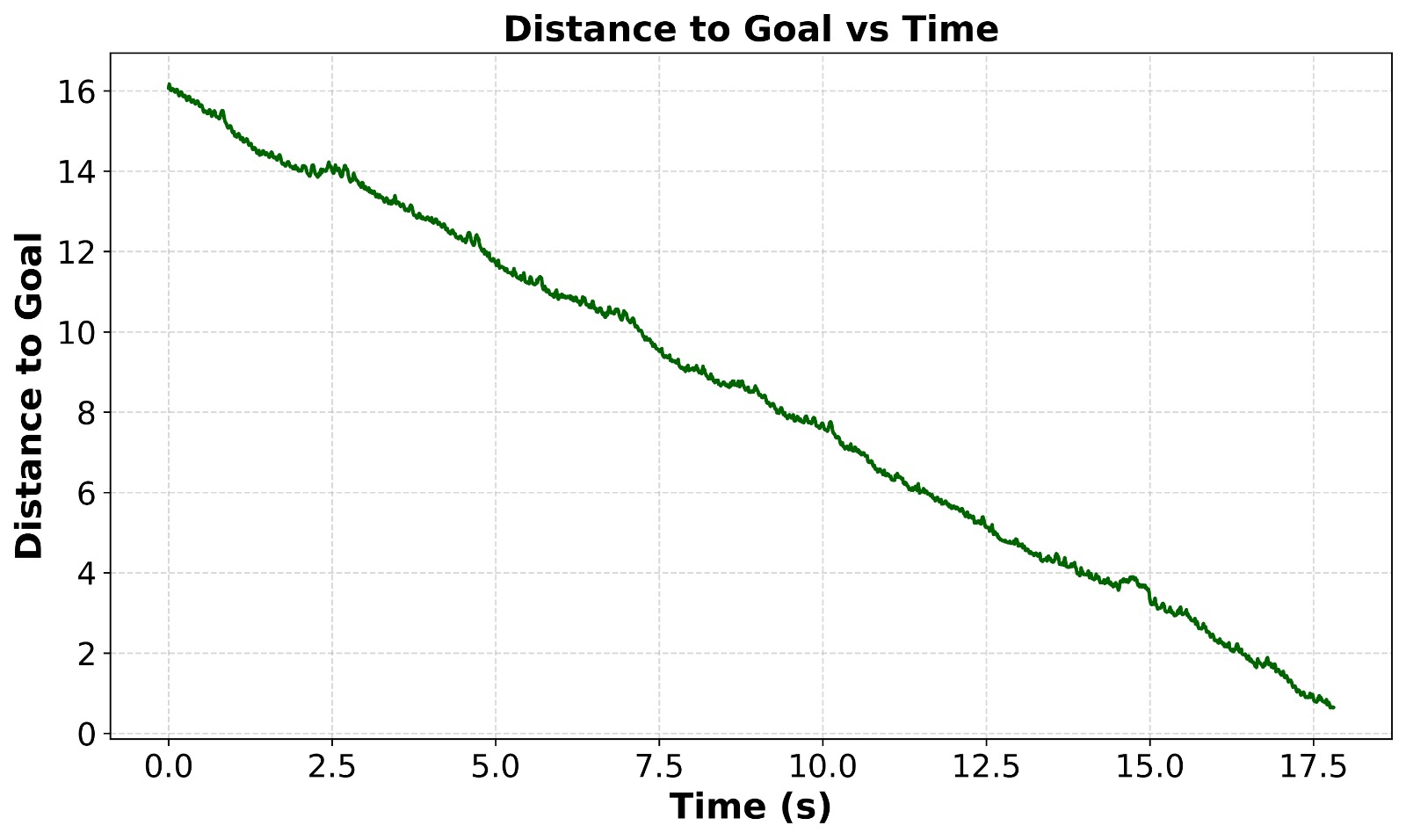} \\
    \end{tabular}
    \caption{Performance metrics of the NEAT controller. The plots show (top row, left to right) the evolution of the \textit{x}-position, \textit{y}-position for NEAT, CBRL \cite{Jia2021}, MFRL \cite{8103164}, DLDC \cite{Sartoretti_2019}, and head angle of NEAT over time, and (bottom row, left to right) control frequency $\omega$, offset $\phi_0$, and distance to goal with time of the NEAT controller.}
    \label{fig:dynamics}
\end{figure*}

\begin{table}[t]
\centering
\caption{Quantitative Analysis of Different Methods}
\label{tab:quantitative}
\begin{tabular}{lcccc}
\hline
\textbf{Metric} & \textbf{NEAT} & \textbf{CBRL\cite{Jia2021}} & \textbf{MFRL\cite{8103164}} & \textbf{DLDC\cite{Sartoretti_2019}}\\
\hline
Routing Time (s) & \textbf{16.64\textsuperscript{\textdaggerdbl}} & 23.10 & 52.41 & 52.95 \\
Collisions       & 3            & \textbf{2}                  & 10             & 8     \\
Controller Size (KB) & \textbf{19.12} & 71.51                       & -              & -     \\
Number of Params   & \textbf{4897}  & 18308                       & -              & -     \\
\hline
\end{tabular}

\vspace{0.5em} 
\small 
\textsuperscript{\textdaggerdbl}Normalized to a step time of 10ms.
\end{table}


\begin{figure*}[t]
    \centering
    \begin{tabular}{ccc}
        \includegraphics[width=0.3\textwidth, height=3cm]{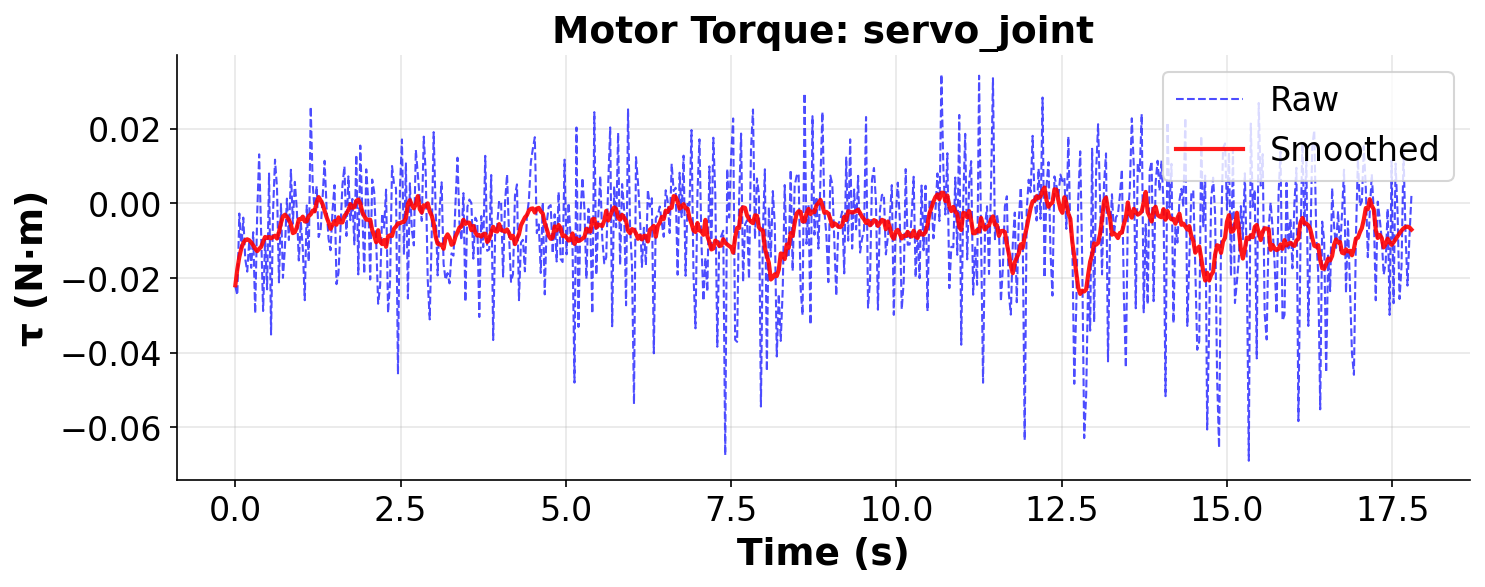} & 
        \includegraphics[width=0.3\textwidth, height=3cm]{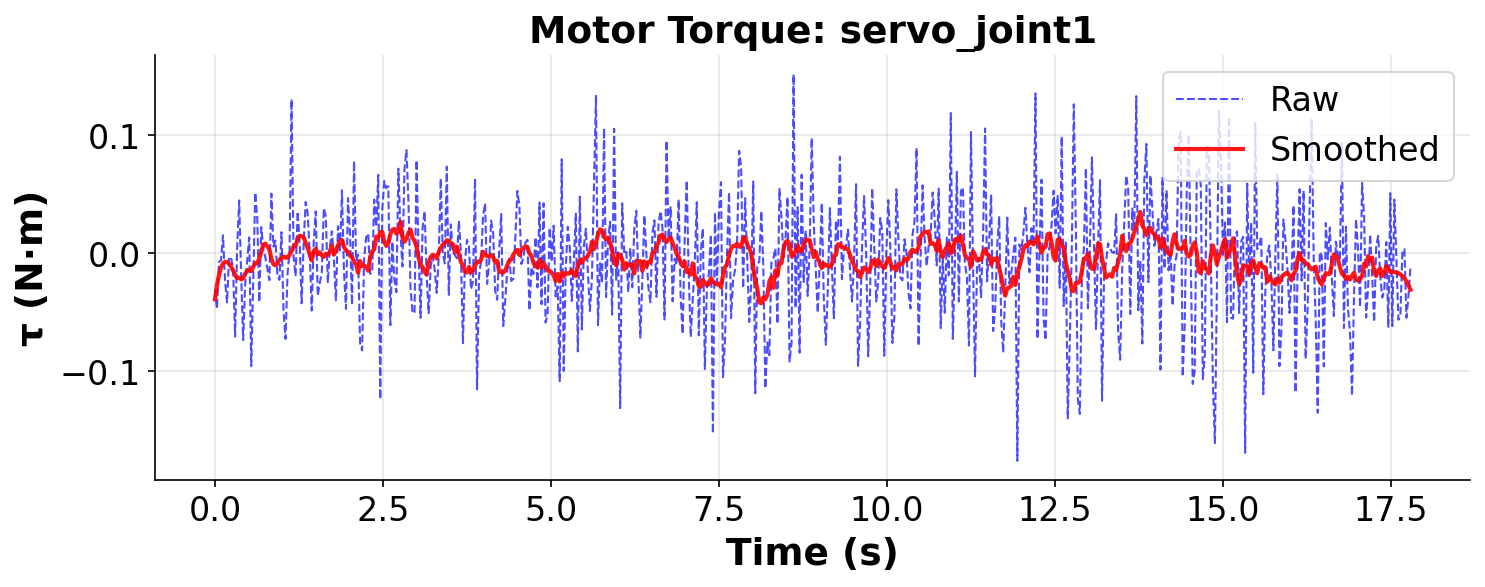} &
        \includegraphics[width=0.3\textwidth, height=3cm]{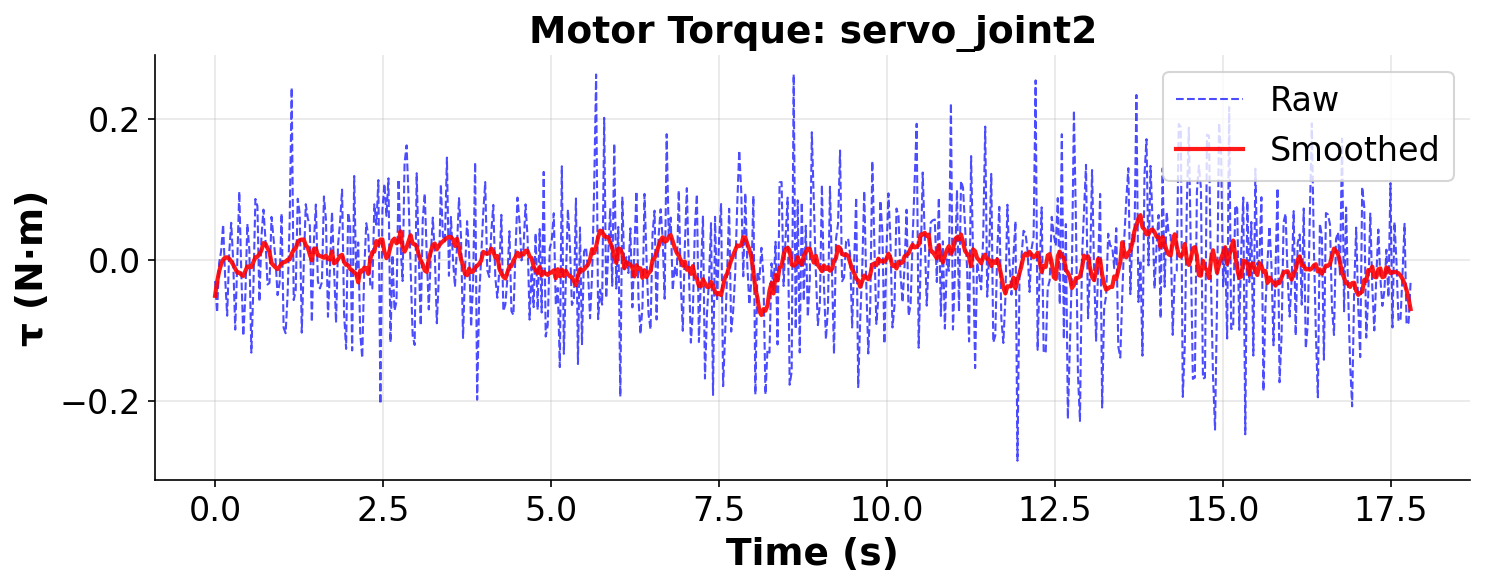} \\
        \includegraphics[width=0.3\textwidth, height=3cm]{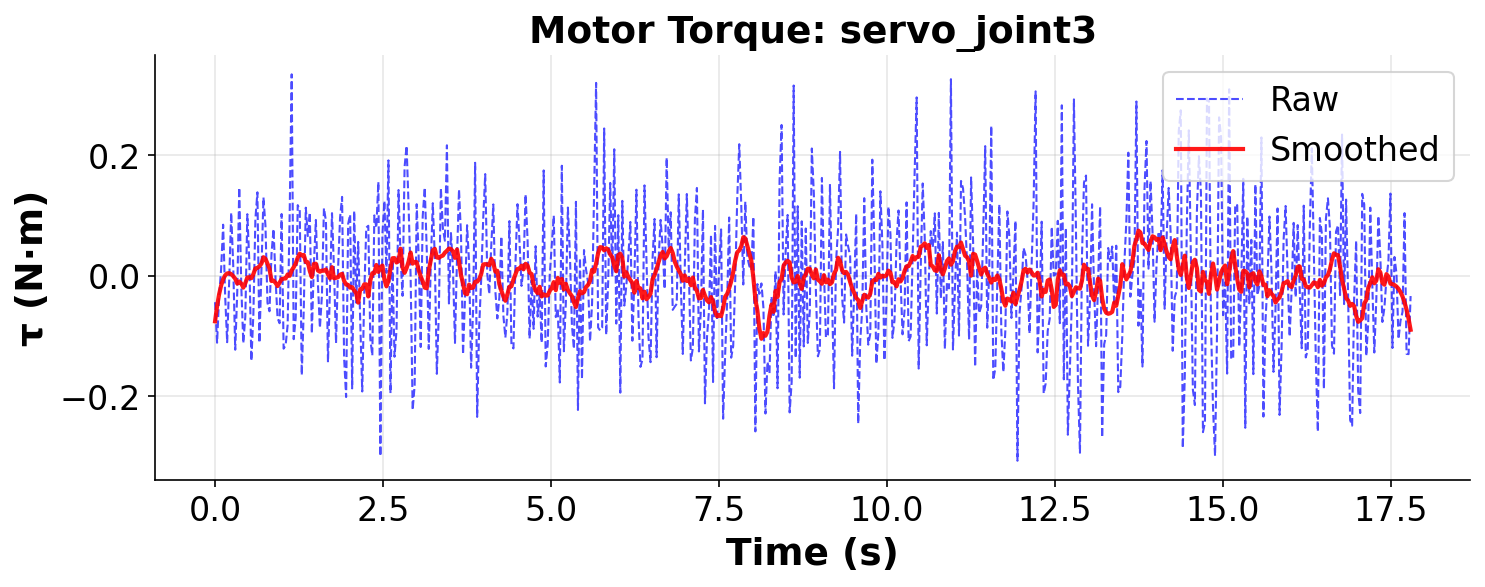} &
        \includegraphics[width=0.3\textwidth, height=3cm]{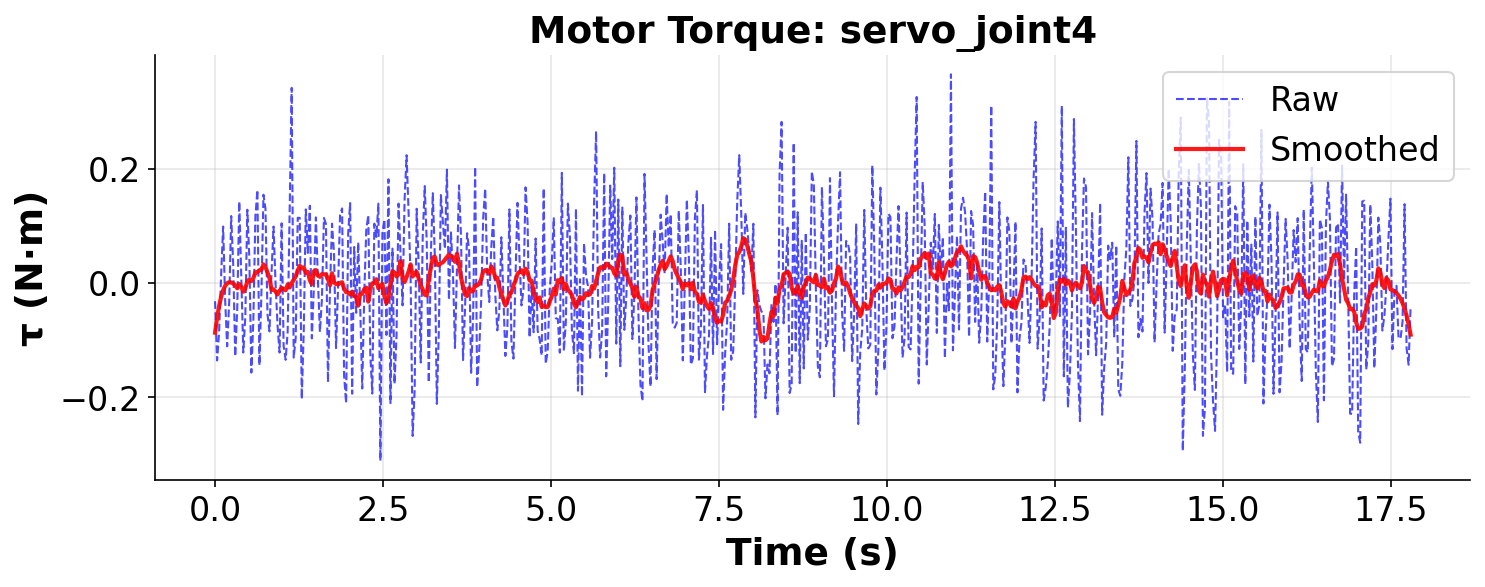} &
        \includegraphics[width=0.3\textwidth, height=3cm]{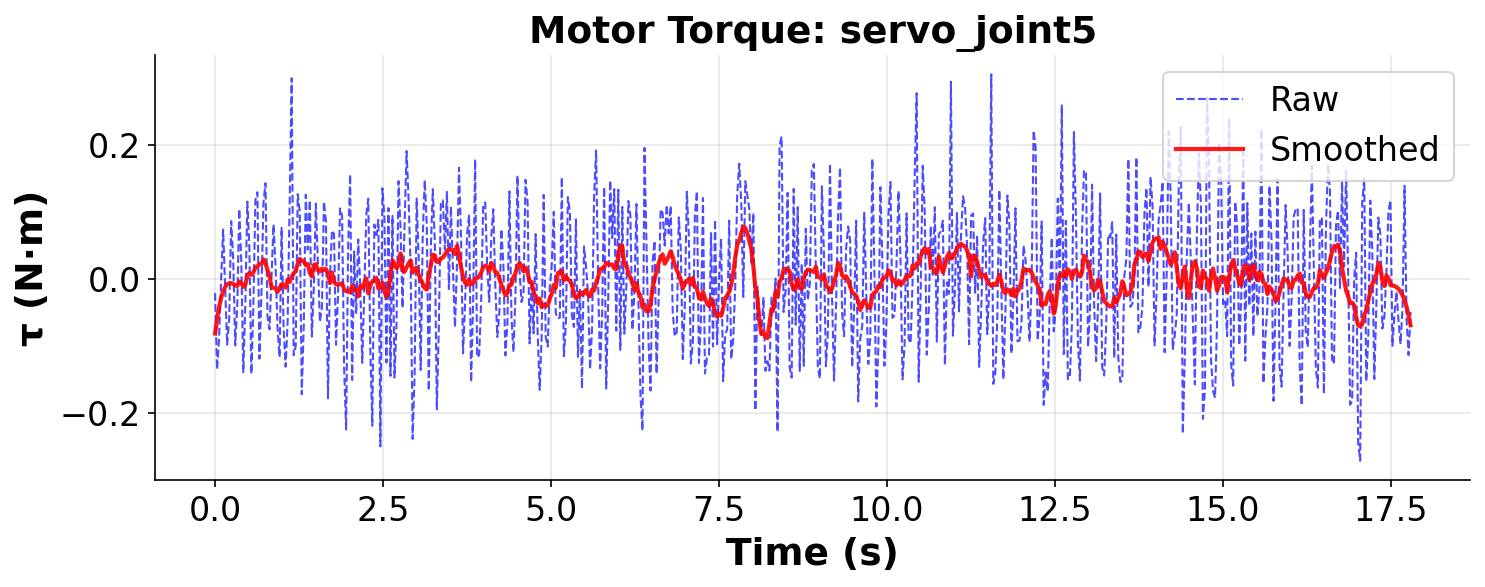} \\
        \multicolumn{3}{c}{%
            \includegraphics[width=0.3\textwidth, height=3cm]{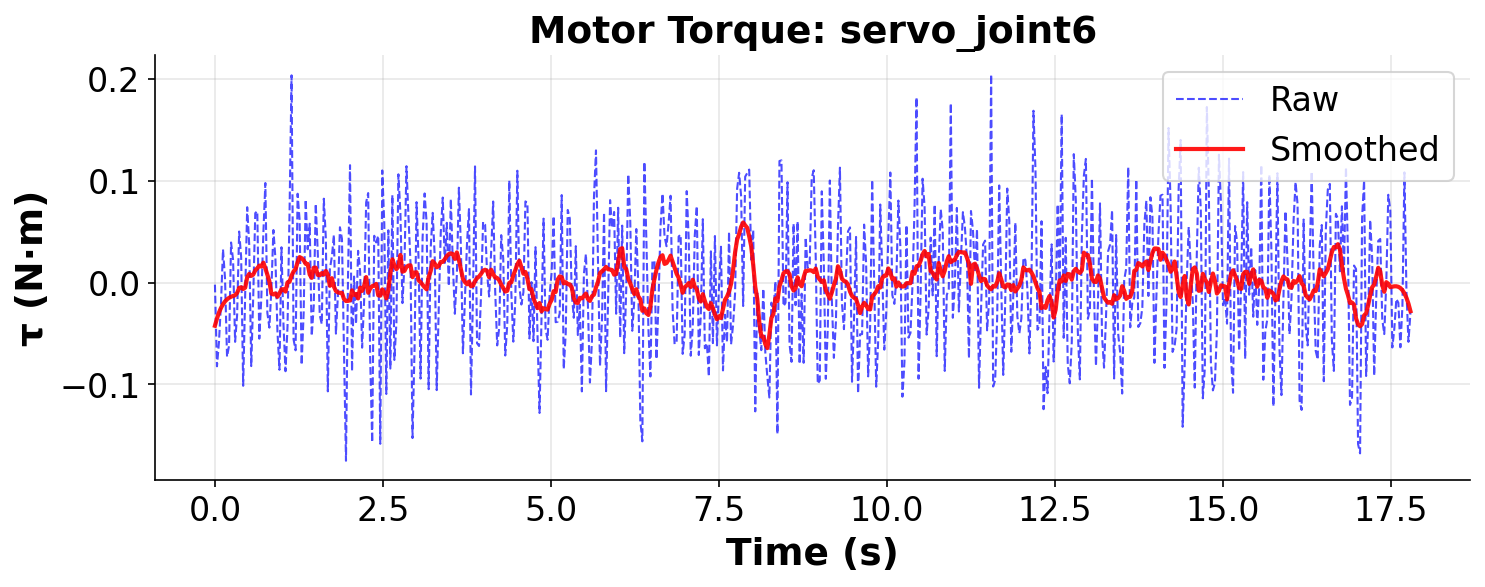} \hspace{1em}
            \includegraphics[width=0.3\textwidth, height=3cm]{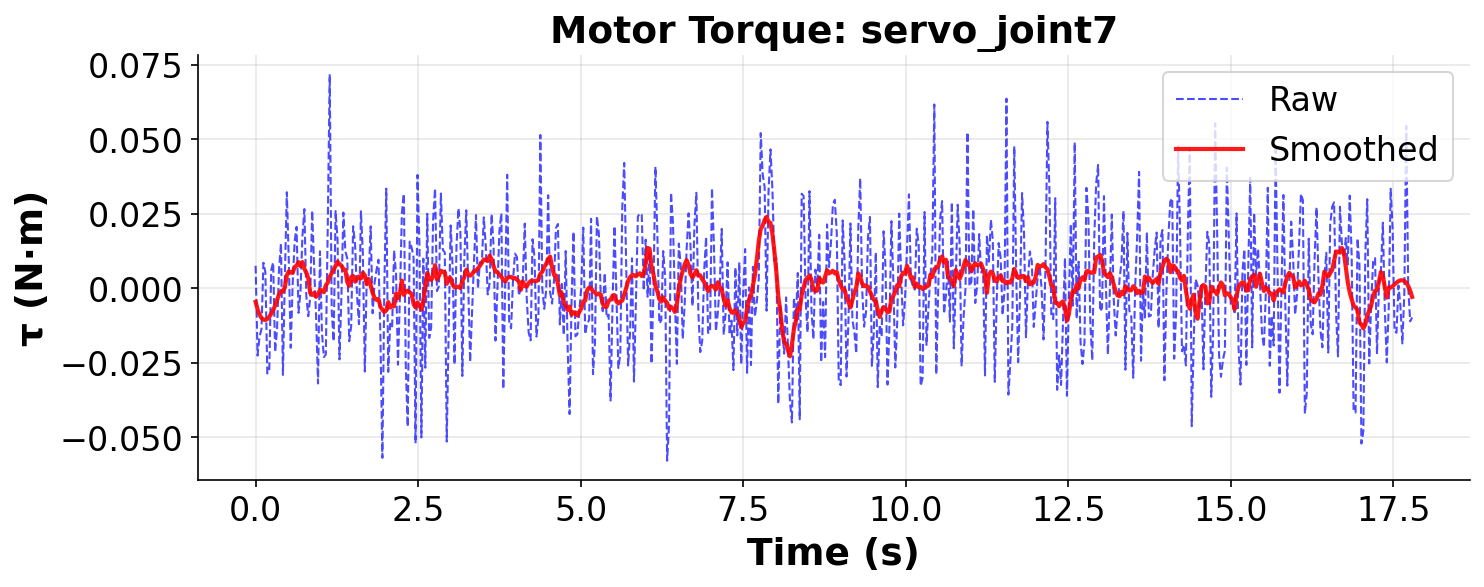}
        } \\
    \end{tabular}
    \caption{Measured torques of all joints of the snake.}
    \label{fig:torques}
\end{figure*}

\begin{figure*}[t]
    \centering
    \begin{tabular}{ccc}
        \includegraphics[width=0.43\textwidth, height=3.7cm]{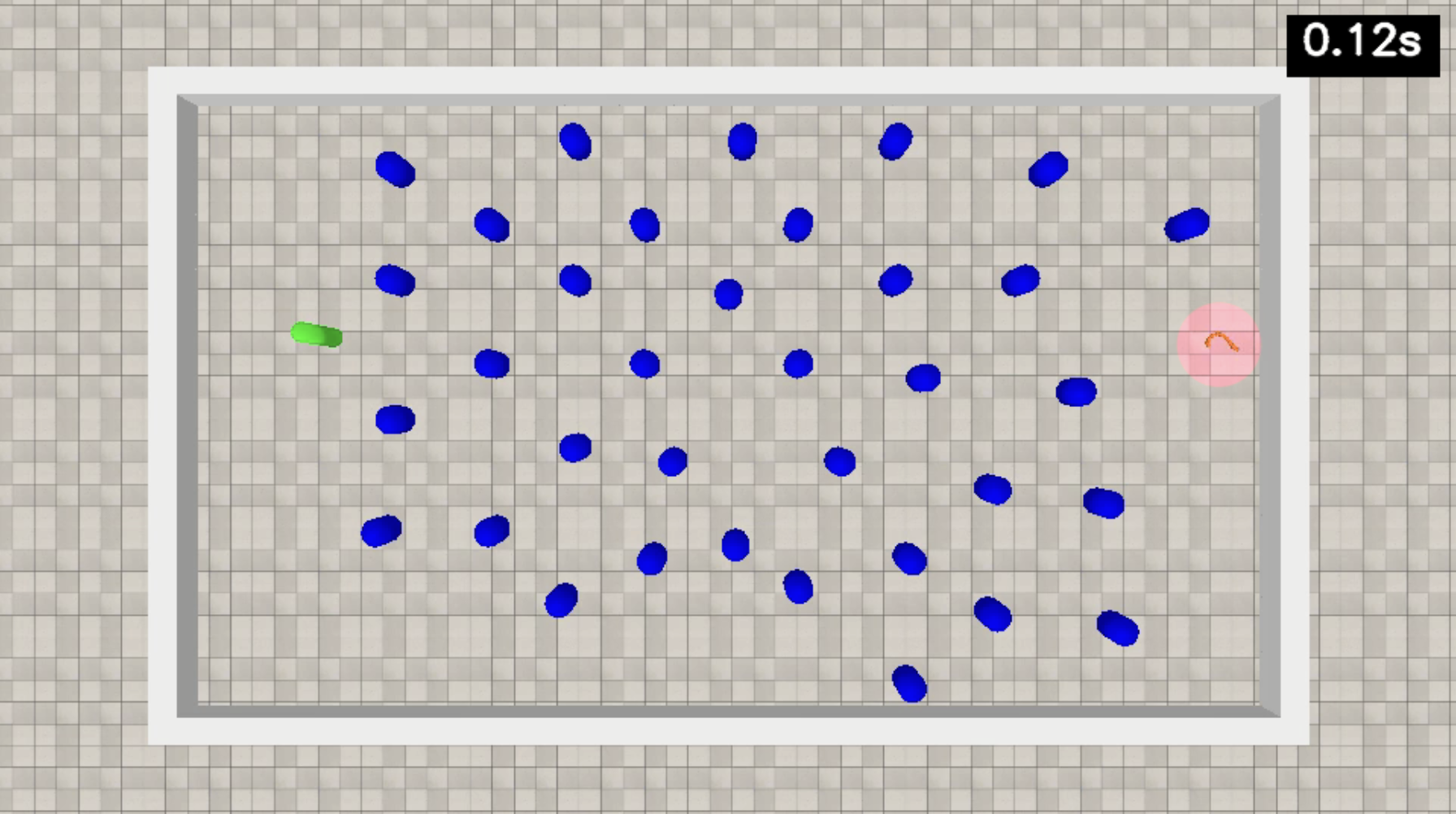} & 
        \includegraphics[width=0.43\textwidth, height=3.7cm]{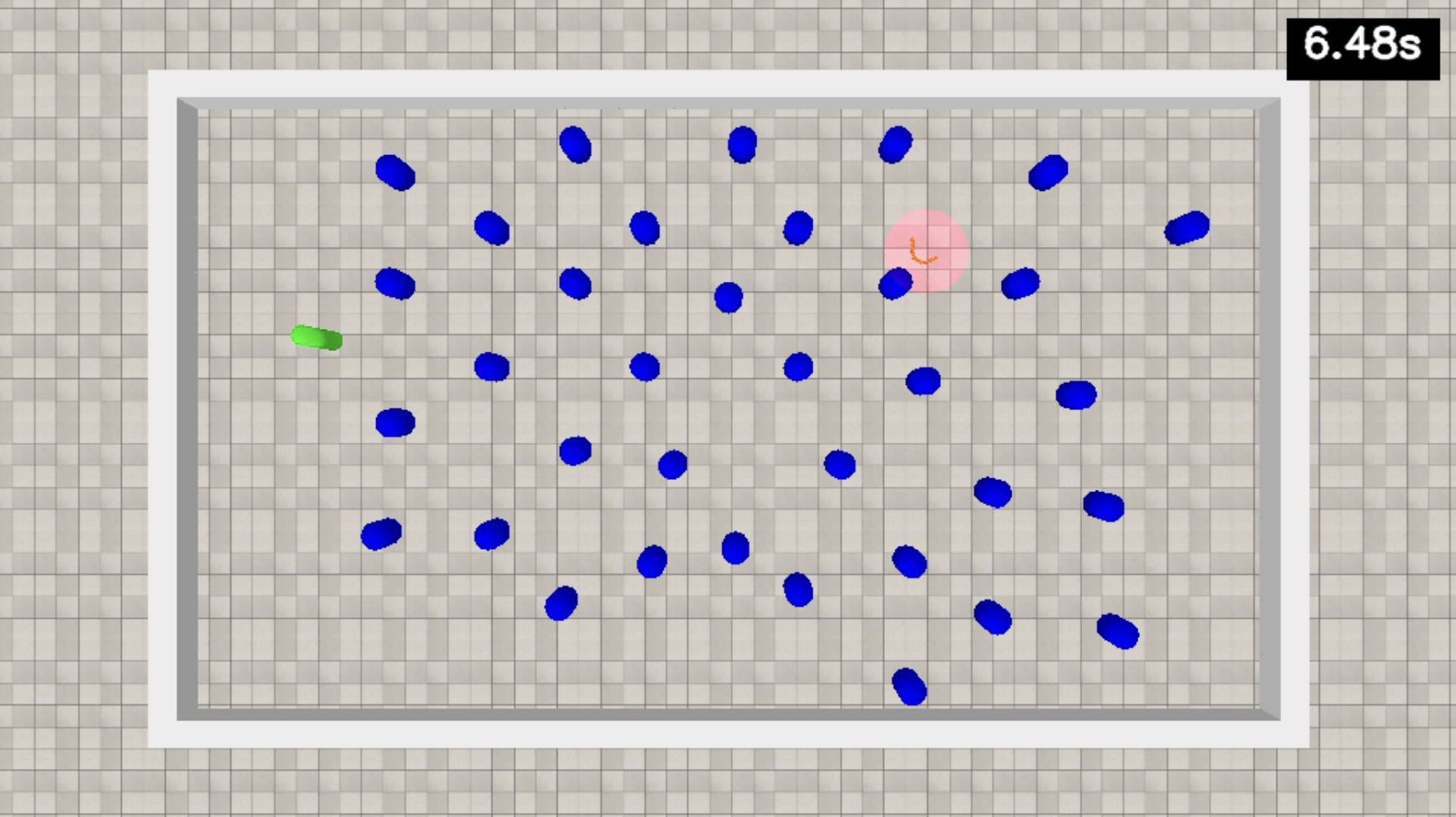} \\
        \includegraphics[width=0.43\textwidth, height=3.7cm]{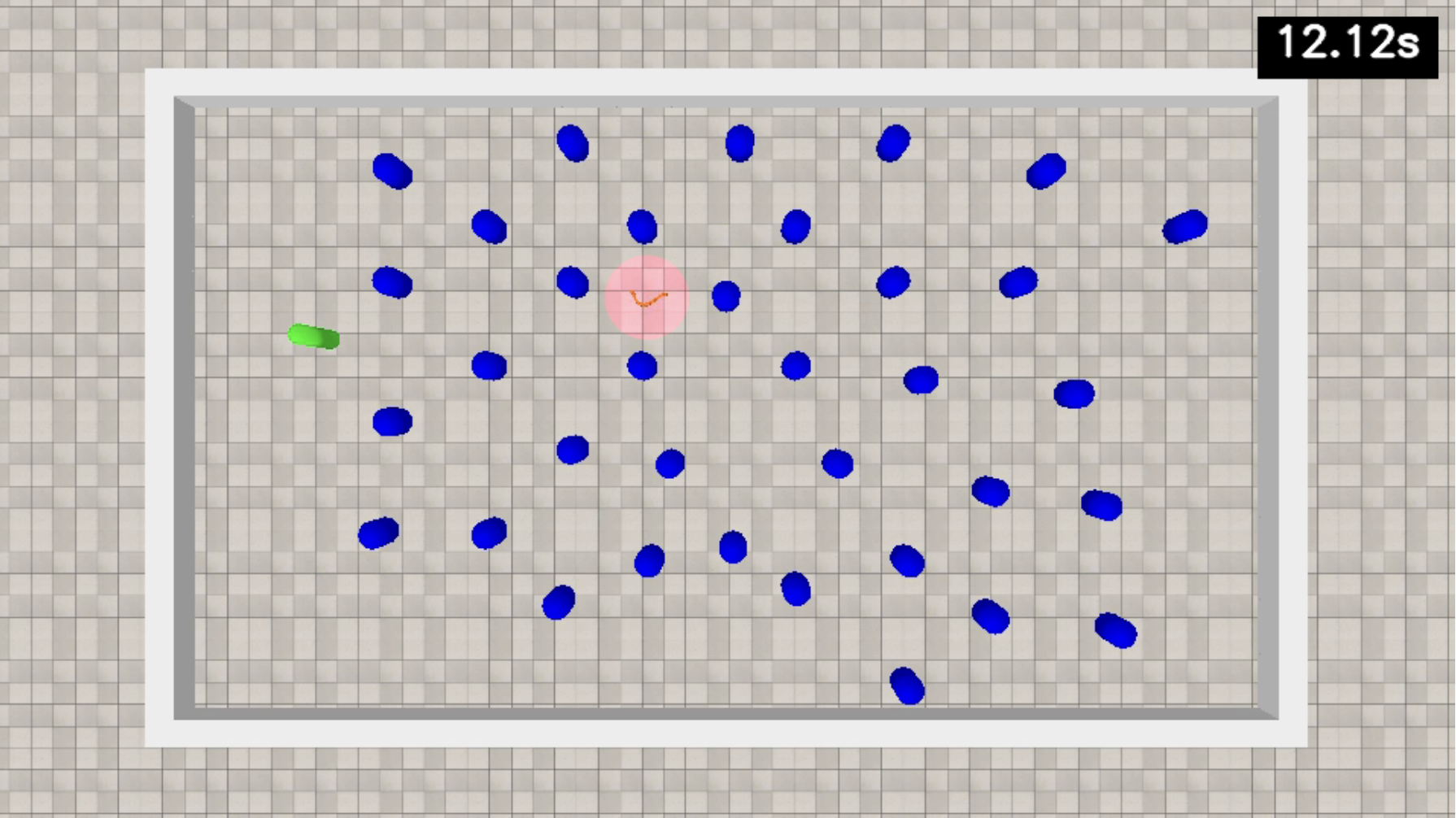} & 
        \includegraphics[width=0.43\textwidth, height=3.7cm]{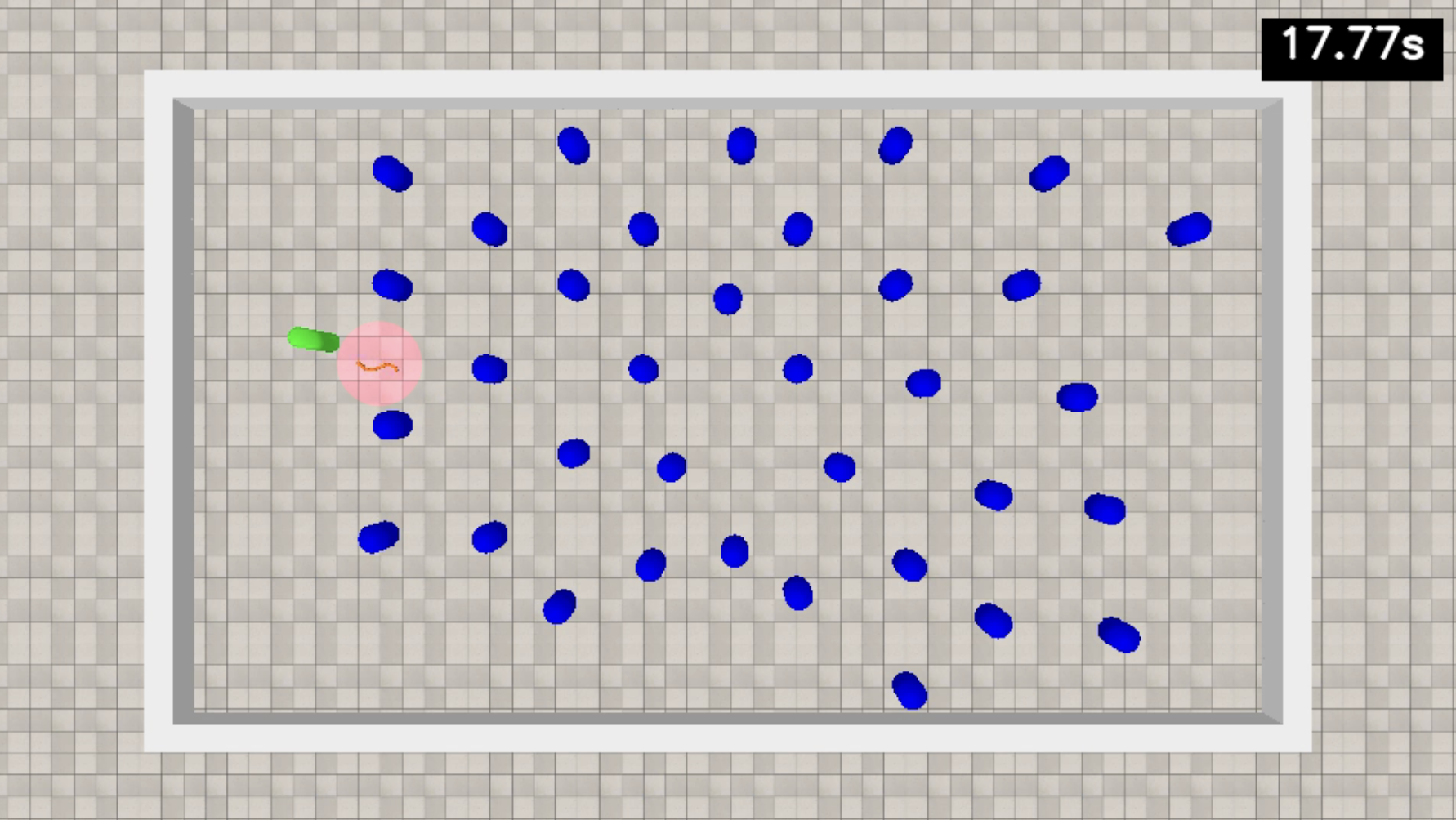} \\
    \end{tabular}
    \caption{Snapshots of the snake navigation (red) at selected timestamps (0.12 s, 6.48 s, 12.12 s, and 17.77 s).}
    \label{fig:snaps}
\end{figure*}

Each training simulation episode is allowed a maximum of $20,000$ steps. To encourage goal-directed behaviour, a reward of $2000$ is given when the snake reaches the goal, while a penalty of $20$ is applied for each obstacle collision (\ref{f_coll_eq}). The gain factor for the progress reward (\ref{f_prog_eq}) is set to $20$. The corridor penalty (\ref{f_corr_eq}) is applied when the snake goes out of the bounds [$-0.5$, $3.5$]. The penalty factor for this is set to $3.0$. Finally, a step-wise bonus (\ref{f_live_eq}) of $0.05$ is applied to prevent premature stagnation. An episode is terminated early if the snake collides more than $6$ times, but up to $6$ collisions are tolerated to allow the robot to still learn from partial trajectories. The time for each step is 83.3ms. During training, the success condition is defined as reaching within 1.5 units of the goal to encourage progress toward the goal without requiring exact convergence. For inference, this tolerance is reduced to 0.5 units for stricter evaluation. We observe that the policy generalizes well under this setting, typically reaching the goal without any additional collisions.

\subsection{Simulation results}
The proposed NEAT methodology was evaluated for its navigation ability and efficiency by comparing it to previous state-of-the-art methods. Fig. \ref{fig:prev results} compares the snake robot's trajectories for the proposed NEAT approach against the CBRL \cite{Jia2021}, MFRL \cite{8103164}, and DLDC \cite{Sartoretti_2019} methods. These trajectories show that the NEAT model is able to navigate the obstacle-dense environment much better than the MFRL and DLDC methods. Specifically, the path generated by NEAT is direct and efficient. In contrast, the paths generated by MFRL and DLDC are much more erratic, suggesting that the controller struggled to find a clear path while avoiding collisions.\\
Table \ref{tab:quantitative} provides an average comparison of the performance metrics, averaged over 5 simulation runs, with the head angle of the snake restricted to $[-45^\circ, 45^\circ]$. The simulation results confirm that the serpentine controller optimized with NEAT generates a significantly more time-optimal trajectory compared to previous state-of-the-art algorithms. To ensure a fair comparison of routing times, we normalized the simulation step time to 10 ms, matching the methodology used in \cite{Jia2021}. The NEAT approach achieves the shortest routing time (16.64s), outperforming CBRL (23.10s) and providing a significant improvement over MFRL (52.41s) and DLDC (52.95s). In terms of obstacle avoidance, NEAT also proved robust, averaging 3 collisions, which is a significant improvement over the 10 and 8 collisions recorded for MFRL and DLDC, respectively.\\
A key contribution of this work is the computational efficiency of the evolved controller. The proposed NEAT controller is extremely compact at only 19.12KB, comprising 4,897 parameters. This is nearly four times smaller than the controller generated by the CBRL method, which requires 71.51KB and 18,308 parameters.  This substantial reduction is a major advantage for practical applications, as it facilitates deployment on hardware with limited computational and memory resources.\\
Fig. \ref{fig:dynamics} details the dynamics of the NEAT controller. The efficiency of the navigation is evident in the plot of distance to the goal, which shows a near-linear decrease, indicating constant and direct progress towards the goal. The plots of head angle and angular frequency $(\omega)$ shows that the evolved controller generates a stable high-frequency oscillatory gait. This continuous actuation allows the snake to maintain momentum and make subtle adjustments while moving, which is key to its smooth navigation around obstacles.
Fig. \ref{fig:torques} illustrates the torque profiles for each of the eight servo motors. The plots show the actuator-level commands produced by the NEAT controller. All the torque values are within a range of $[-0.2, 0.2]$, indicating a highly efficient controller that avoids motor saturation and minimzes energy wastage. This highlights the robust nature of the controller at the actuator level. The motion of the snake robot at different timestamps in the environment is shown in Fig. \ref{fig:snaps}.


\subsection{Discussion}
The experimental results support the central hypothesis, i.e.,  the proposed algorithm can match and even outperform established state-of-the-art techniques. Beyond its performance, the algorithm's design simplicity offers a distinct advantage in computational efficiency. We attribute its significantly reduced memory footprint to this simple structure, which stands in contrast to the more complex architectures of the state-of-the-art methods. This outcome suggests a promising trade-off, where high performance can be achieved without the resource burden of more intricate models. Additionally, such a simple architecture allows one to implement the proposed approach in a real time situation, where the actual robot can find the local obstacle avoiding control solution based on real time LiDAR and sensor data.

\subsection{Ablation Studies}
Ablation studies were performed to analyze the effect of maximum head angle on the performance metrics in terms of number of collisions, number of steps to reach the goal and the final training fitness. The best results of each angle are shown in Table \ref{tab:head_angle_ablation}. The results indicate that $45^\circ$ is the optimal head angle as it was the only setting to achieve zero collisions, while achieving the highest fitness of $2181.1$.

\begin{table}[t] 
\centering
\caption{Ablation Study on Maximum Head Angle Limits}
\label{tab:head_angle_ablation}
\begin{tabular}{c c c c}
\hline
\textbf{Max Head Angle (°)} & \textbf{Collisions} & \textbf{Steps to Goal} & \textbf{Final Fitness} \\
\hline
30  & 4  & 1523   & 2145.6 \\
45  & 0 & 1781 & 2181.1 \\
60  & 3 & 2203 & 1872.7 \\
90 & 2 & 1927 & 1812.2 \\
180 & 4  & 2046 & 1730.7 \\
\hline
\end{tabular}
\end{table}

We observed that a more restrictive head angle of $30^\circ$ resulted in more collisions, indicating that the narrow field of motion hindered the controller's ability to avoid obstacles effictively. Conversely, increasing the maximum head angle beyond $45^\circ$ led to a consistent increase in collisions and steps, and a lower final fitness. This suggests that the excessive, uncontrolled head movement results in instability.

\section{Conclusion} \label{sec:conc}
This work presents a NEAT based obstacle avoiding tracking control of a planar snake robot in an obstacle dense environment to reach the intended goal with time efficiency and minimum number of collisions. NEAT employs joint angles, link positions, head link position as well as obstacle positions in the vicinity as inputs and generates the frequency and offset angle of the serpenoid gait function of the robot to dynamically control the motion of the snake robot in the environment. A reward function has been proposed to minimize the number of collisions with obstacles as well as the distance between the head link and the goal and the time required by the robot to reach the goal. The proposed approach employs a single neural network to find a near optimal path iteratively, making it computationally efficient compared to the DQN based approaches as in the state-of-the-art. The initialization parameters and the subsequent iterative selection has to be suitably chosen as, the single neural network based NEAT is prone to local convergence. The proposed framework has been verified through physics engine simulation on PyBullet which shows comparable result to the CBRL based methodology and even improves on the other state-of-the-art methodologies by minimizing collisions with significantly lower computational requirement. The future work will be aimed at exploring other advanced versions and real-time implementation of NEAT.










\bibliographystyle{IEEEtran}  
\bibliography{IEEEabrv, bibieee} 

\end{document}